\documentclass{article}

\usepackage{microtype}
\usepackage[title]{appendix}
\usepackage{graphicx}
\usepackage{amsmath}
\usepackage{bm}
\usepackage{placeins}
\usepackage{amsfonts}
\usepackage{amsthm}
\usepackage{caption}
\usepackage{subcaption}
\usepackage{subfiles}
\usepackage{booktabs} 
\usepackage{siunitx}
\usepackage{comment}

\usepackage{xr-hyper}
\usepackage{hyperref}
\makeatletter
\newcommand*{\addFileDependency}[1]{
  \typeout{(#1)}
  \@addtofilelist{#1}
  \IfFileExists{#1}{}{\typeout{No file #1.}}
}
\makeatother

\newtheorem{theorem}{Theorem}[section]

\newtheorem{proposition}[theorem]{Proposition}
\theoremstyle{definition}
\newtheorem{definition}{Definition}[section]

\usepackage[accepted]{icml 2020}

\icmltitlerunning{Mix-n-Match Calibration}
\usepackage{subfiles} 

\begin{document}

\twocolumn[
\icmltitle{{\em Mix-n-Match}: Ensemble and Compositional Methods for Uncertainty Calibration in Deep Learning}


\begin{icmlauthorlist}
\icmlauthor{Jize Zhang}{to}
\icmlauthor{Bhavya Kailkhura}{to}
\icmlauthor{T. Yong-Jin Han}{to}
\end{icmlauthorlist}

\icmlaffiliation{to}{Lawrence Livermore National Laboratories Livermore, CA 94550}

\icmlcorrespondingauthor{Jize Zhang}{zhang64@llnl.gov}

\icmlkeywords{Deep learning, confidence estimates, calibration, error evaluation, Ensemble}

\vskip 0.3in
]



\printAffiliationsAndNotice{}  

\begin{abstract}
This paper studies the problem of post-hoc calibration of machine learning classifiers. We introduce the following desiderata for uncertainty calibration: (a) accuracy-preserving, (b) data-efficient, and (c) high expressive power. We show that none of the existing methods satisfy all three requirements, and demonstrate how {\em Mix-n-Match} calibration strategies (i.e., ensemble and composition) can help achieve remarkably better data-efficiency and expressive power while provably maintaining the classification accuracy of the original classifier. 
{\em Mix-n-Match} strategies are generic in the sense that they can be used to improve the performance of any off-the-shelf calibrator.
We also reveal potential issues in standard evaluation practices.
Popular approaches (e.g., histogram-based expected calibration error (ECE)) may provide misleading results especially in small-data regime.
Therefore, we propose an alternative data-efficient kernel density-based estimator for a reliable evaluation of the calibration performance and prove its asymptotically unbiasedness and consistency. 
Our approaches outperform state-of-the-art solutions on both the calibration as well as the evaluation tasks in most of the experimental settings.
Our codes are available at \href{https://github.com/zhang64-llnl/Mix-n-Match-Calibration}{https://github.com/zhang64-llnl/Mix-n-Match-Calibration}.



\end{abstract}

\section{Introduction}

\begin{figure}[!t]
  \centering
  \begin{subfigure}{0.5\textwidth}
    \includegraphics[clip,width=1\columnwidth]{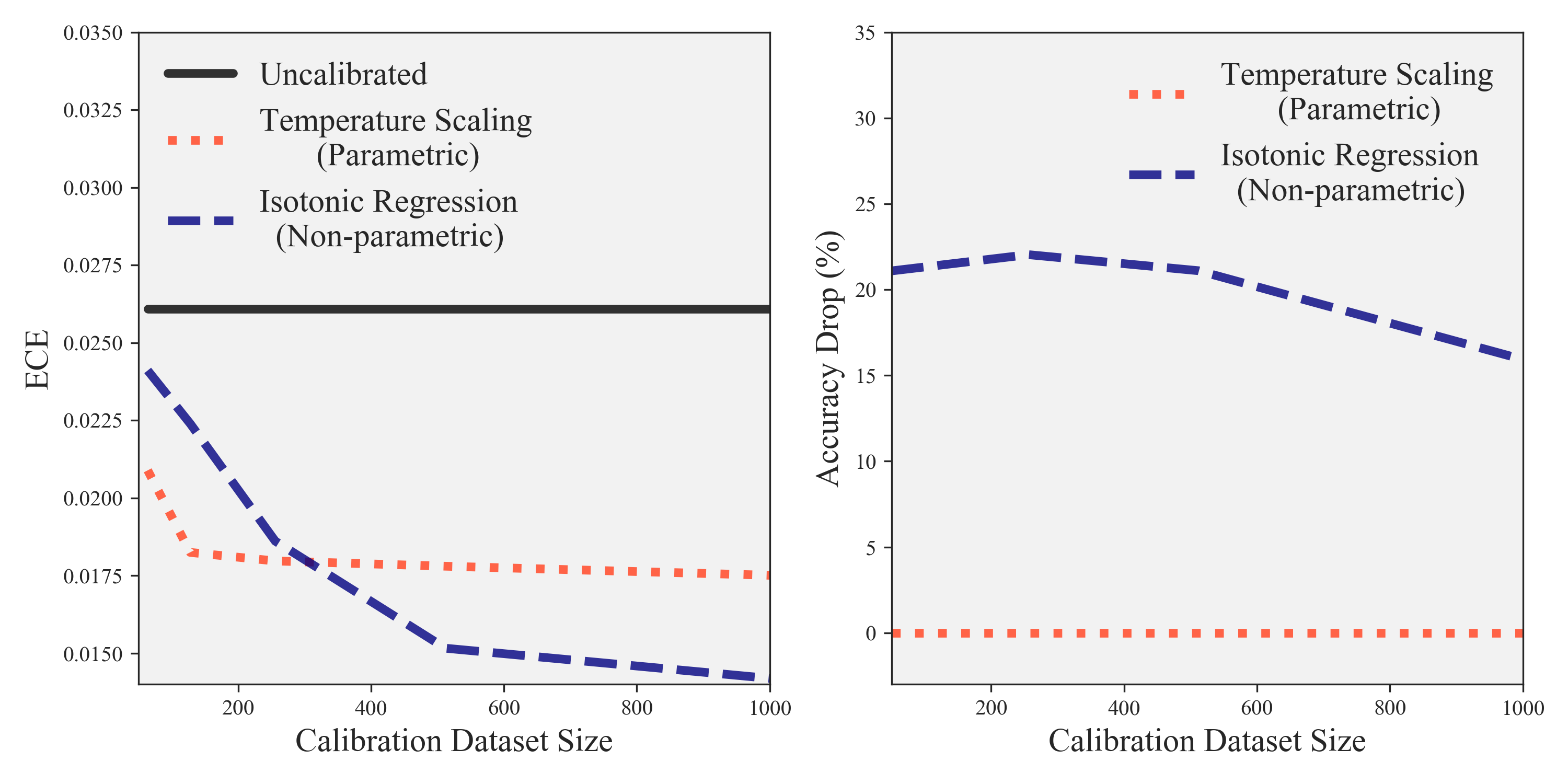}
  \end{subfigure}
   \begin{subfigure}{0.5\textwidth}
  \includegraphics[clip,width=1\columnwidth]{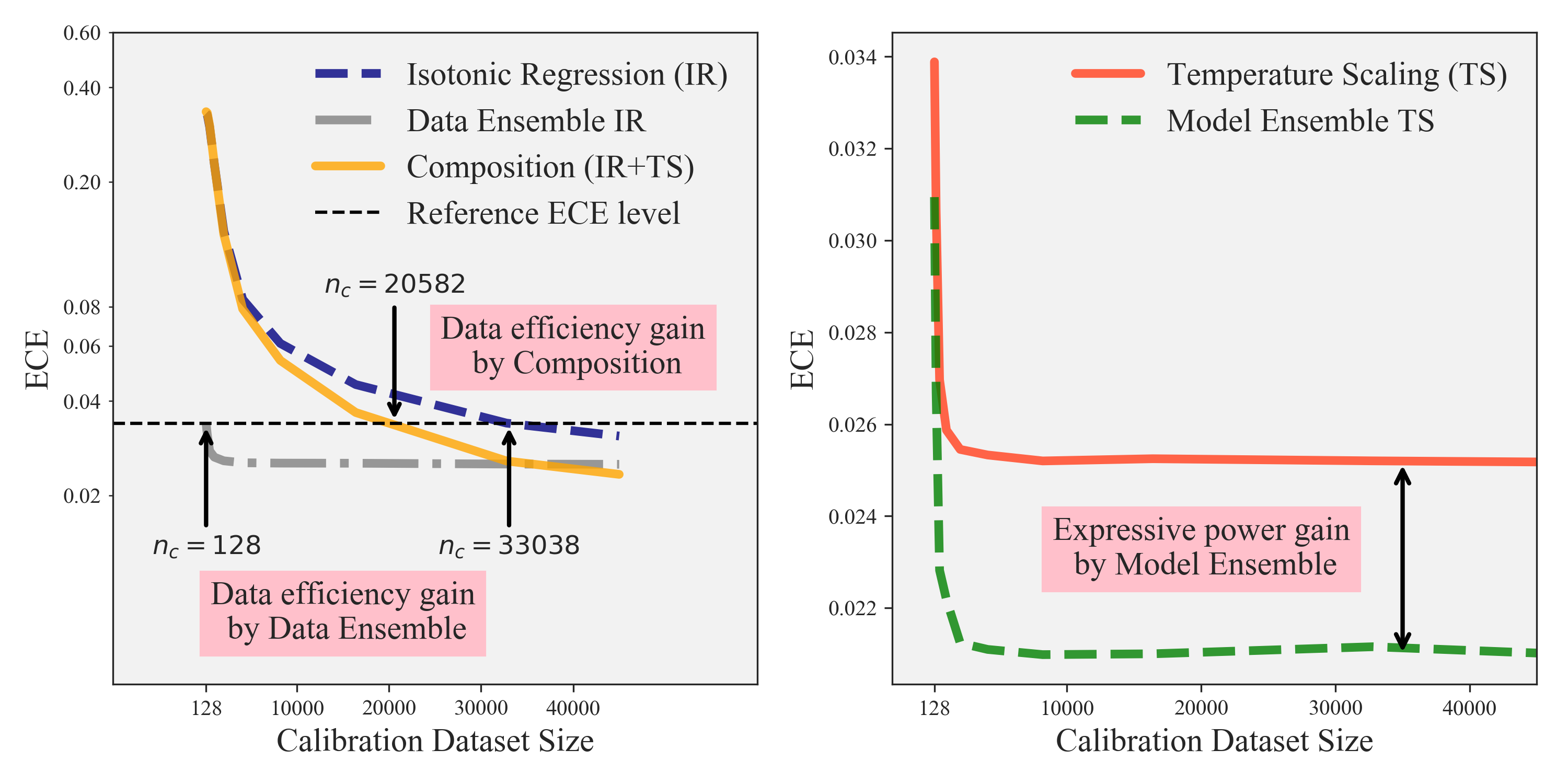}
  \end{subfigure}
  \vspace{-0.12in}
  \caption{(Top): (left) Temperature Scaling (TS) \cite{guo2017calibration} is data-efficient (initial rapid ECE drop) but not expressive (fails to make progress later); in contrast, Isotonic Regression (IR) \cite{zadrozny2002transforming} is more expressive but data-inefficient; (right) IR does not preserve accuracy and introduces significant accuracy drop. (Bottom) {\em Mix-n-Match}: (left) Data Ensemble and Composition improve the data efficiency of IR, and (right) Model Ensemble enhances the expressive power of TS. All results are for calibrating a 50-layer Wide ResNet on ImageNet, apart from (a) left (28-layer Wide ResNet on CIFAR-10).}
  \label{fig:summary}
  \vspace{-0.1in}
\end{figure}

Machine learning (ML) models, e.g., deep neural networks, are increasingly used for making potentially important decisions in applications ranging from object detection \cite{girshick2015fast}, autonomous driving \cite{chen2015deepdriving} to medical diagnosis \cite{litjens2017survey}. 
Several of these applications are high-regret in nature and incorrect decisions have significant costs. Therefore, besides achieving high accuracy, it is also crucial to obtain reliable uncertainty estimates, which can help deciding whether the model predictions can be trusted \cite{jiang2018trust,kendall2017uncertainties}. 
Specifically, a classifier should provide a calibrated uncertainty measure in addition to its prediction. A classifier is well-calibrated, if the probability associated with the predicted class label matches the probability of such prediction being correct~\cite{brocker2009reliability,dawid1982well}. Unfortunately, many off-the-shelf ML models are not well calibrated \cite{niculescu-Mizil2005predicting,zadrozny2001learning,zadrozny2002transforming}. Poor-calibration is particularly prominent in highly complex models such as deep neural network classifiers  \cite{guo2017calibration,hein2019relu,lakshminarayanan2017simple,nguyen2015deep}. 

A popular calibration approach is to learn a transformation (referred to as a calibration map) of the trained classifier's predictions on a calibration dataset in a {\em post-hoc} manner. Pioneering work along this direction include Platt scaling \cite{platt2000probabilistic}, histogram binning \cite{zadrozny2001learning}, isotonic regression \cite{zadrozny2002transforming}, Bayesian binning into quantiles \cite{naeini2015obtaining}. Recently, calibration methods for multi-class deep neural network classifiers have been developed, which include: temperature, vector \& matrix scaling \cite{guo2017calibration}, Dirichlet scaling \cite{kull2019beyond}, intra order-preserving method \cite{rahimi2020intra} and Gaussian processes based calibration methods \cite{milios2018dirichlet, wenger2019non}.
Besides post-hoc calibrations, there also exist approaches for training ab-initio well calibrated models \cite{kumar2018trainable,lakshminarayanan2017simple,pereyra2017regularizing,seo2019learning,tran2019calibrating}, or representing the prediction uncertainty in a Bayesian framework \cite{blundell2015weight,gal2016dropout,maddox2019simple}.


Ideally, an  uncertainty calibration method should satisfy the following properties: (a) {\em accuracy-preserving} -- calibration process should not degrade the classification accuracy of the original classifier, (b) {\em data-efficiency} -- the ability to achieve well-calibration without requiring a large amount of calibration data, and (c) {\em high expressive power} -- sufficient representation power to approximate the canonical calibration function given enough calibration data.
Despite the popularity of post-hoc calibration, we found that none of the existing methods satisfy all requirements simultaneously (\autoref{fig:summary}).
Yet given practical constraints, such as high data collection costs, high complexity of calibration tasks, and need for accurate classifiers, the development of calibration methods which satisfy all three requirements simultaneously is crucial for the success of real-world ML systems.

After calibrating a classifier, the next equally important step is to reliably evaluate the calibration performance. Most of the existing works judge the calibration performance by the expected calibration error\footnote{Alternative choices also exist, such as the max calibration error \cite{naeini2015obtaining} or the reproducing kernel Hilbert space based calibration measure \cite{widmann2019calibration}. 
} (ECE)~\cite{naeini2015obtaining}. ECE is usually estimated from a reliability diagram and its associated confidence histogram~\cite{guo2017calibration,naeini2015obtaining}. However, histogram-based ECE estimators can be unreliable (e.g., asymptotically biased or noisy) due to their sensitivity to binning schemes \cite{ashukha2019pitfalls,ding2020revisiting,nixon2019measuring, vaicenavicius2019evaluating}. Additionally, as a density estimator, histogram is known to be less data-efficient than alternative choices, such as kernel density estimators \cite{scott1992multivariate}.
Therefore, it is of utmost importance to develop reliable and data-efficient methods to evaluate the calibration performance.

To achieve the aforementioned objectives, this paper makes the following contributions: 
\begin{enumerate}
    \item We introduce the following desiderata for uncertainty calibration -- (a) accuracy-preserving, (b) data-efficient, and (c) expressive.
    \item We propose ensemble and compositional calibration strategies to achieve high data-efficiency and expressive power while provably preserving accuracy.
    \item We propose a data-efficient kernel density estimator for a reliable evaluation of the calibration performance.
    \item Using extensive experiments, we show that the proposed {\em Mix-n-Match} calibration schemes achieve remarkably better data-efficiency and expressivity upon existing methods while provably preserve accuracy.
\end{enumerate}

\section{Definitions and Desiderata}

Consider a multi-class classification problem. The random variable $X \in \mathcal{X}$ represents the input feature, and $Y=(Y_1,\ldots,Y_L) \in \mathcal{Y}$ represents the $L$-class one-hot encoded label. Let $f: \mathcal{X}\rightarrow \mathcal{Z} \subseteq \Delta^L$ be a probabilistic classifier that outputs a prediction probability (or confidence) vector $z=f(x)=(f_1(x),\ldots,f_L(x))$, where $\Delta^L$ is  the probability simplex $\{(z_1,\ldots,z_L)\in[0,1]^L|\sum_{l=1}^{L}z_l=1\}$. Let $\mathbb{P}(Z,Y)$ denote the joint distribution of the prediction $Z$ and label $Y$. Expectations ($\mathbb{E}$) are taken over this distribution unless otherwise specified. Let the {\em canonical calibration function} $\pi(z)$ represents the actual class probability conditioned on the prediction $z$ \cite{vaicenavicius2019evaluating}:
\begin{equation}
\label{pi}
\begin{split}
    \pi(z)&=(\pi_1(z),\ldots,\pi_L(z))\\
    \text{with }&\pi_l(z)=\mathbb{P}[Y_l=1|f(X)=z].       
\end{split}
\end{equation}

We would like the predictions to be calibrated, which intuitively means that it represents a true probability.
The formal definition of calibration is as follows:
\theoremstyle{definition}
\begin{definition}
\label{perfect}
The classifier $f$ is {\em perfectly calibrated}, if for any input instances $x \in \mathcal{X}$, the prediction and the canonical calibration probabilities match: $z=\pi(z)$ \cite{dawid1982well}.
\end{definition}

We focus on a post-hoc approach for calibrating a pre-trained classifier, which consists of two steps: $(1)$ finding a calibration map $T: \Delta^L \rightarrow \Delta^L$ that adjusts the output of an existing classifier to be better calibrated, based on a set of $n_c$ calibration data samples; and $(2)$ evaluate the calibration performance based on a set of $n_e$ evaluation data samples. 
Next, we discuss both steps in detail and highlight shortcomings of current methods. 


\subsection{Calibration Step} 
The first task in the calibration pipeline is to learn a calibration map $T$ based on $n_{c}$ calibration data samples $\{(z^{(i)},y^{(i)})\}_{i=1}^{n_c}$. Existing calibration methods can be categorized into two groups:

\textbf{Parametric methods} assume that the calibration map belongs to a finite-dimensional parametric family $\mathcal{T}=\{T(z;\theta)|\theta\in\Theta\ \subseteq \mathbb{R}^{M}\}$. As an example, for binary classification problems, {\em Platt scaling} \cite{platt2000probabilistic}
uses the logistic transformation to modify the prediction probability of a class 
(assuming $z_1$): $T(z_1;a,b) = (1+\exp{(-az_1-b)})^{-1}$, where the scalar parameters $a$, $b$ are learned by minimizing the negative log likelihood on the calibration data set. Parametric methods are easily extendable to multi-class problems, such as {\em temperature, matrix scaling} \cite{guo2017calibration} and {\em Dirichlet scaling} \cite{kull2019beyond}. 

\textbf{Non-parametric methods} assume that the calibration map is described with infinite-dimensional parameters. For binary classification problems, popular methods include: {\em histogram binning} \cite{zadrozny2001learning} which leverages histograms to estimate the calibration probabilities $\pi(z)$ as the calibrated prediction $T(z)$, {\em Bayesian Binning} \cite{naeini2015obtaining} performs Bayesian averaging to ensemble multiple histogram binning calibration maps, and {\em isotonic regression} \cite{zadrozny2002transforming} learns a piecewise constant isotonic function that minimizes the residual between the calibrated prediction and the labels. A common way to extend these methods to a multi-class setting is to decompose the problem as $L$ one-versus-all problems \cite{zadrozny2002transforming}, separately identify the calibration map $T_l$ for each class probability ($z_l$) in the binary manner, and finally normalize the calibrated predictions into $\Delta^L$.

While there are existing approaches tailored for calibrating multi-class deep neural network models, none of them simultaneously satisfy all three proposed desiderata ({\em accuracy-preserving, data-efficient, expressive}). \autoref{fig:summary} (top right) highlights that good calibration capability might come at the cost of classification accuracy for approaches such as isotonic regression. This motivates us to design provably accuracy-preserving calibration methods. Furthermore, the effectiveness of calibration method changes with the amount of calibration data. Parametric approaches are usually data-efficient but have very limited expressive power. On the other hand, non-parametric approaches are expressive but highly data-inefficient. Therefore, in \autoref{fig:summary} (top left), we see that temperature scaling is the best calibration method in data-limited regime, while isotonic regression is superior in data-rich regime. It is thus naturally desirable to design a calibrator that is effective in both data-limited and data-rich regime. However, earlier studies examined calibration methods with fixed dataset size \cite{guo2017calibration,kull2019beyond,wenger2019non}, and shed no light on this issue.

\subsection{Calibration Error Evaluation Step}
The next task in the calibration pipeline is to evaluate the calibration performance based on $n_{e}$ evaluation data points $\{(z^{(i)},y^{(i)})\}_{i=1}^{n_e}$. A commonly used statistics is the expected deviation from $z$ to $\pi(z)$, also called {\em expected calibration error} \cite{naeini2015obtaining}:
\begin{equation}
\label{cali}
\text{ECE}^d(f) = \mathbb{E}\lVert Z-\pi(Z)\rVert_d^d=\int \lVert z-\pi(z)\rVert_d^d \;p(z)\mathop{dz}
\end{equation}
where $\lVert\cdot\rVert_d^d$ denotes the $d$-th power of the $\ell_d$ norm, and $p(z)$ represents the marginal density function of $Z=f(X)$. The original ECE definition adopts $d=1$ \cite{guo2017calibration,naeini2015obtaining}, while $d=2$ is also commonly used \cite{brocker2009reliability,hendrycks2018deep,kumar2019verified}.

Note that probabilities in Eq.~\eqref{pi} and Eq.~\eqref{cali} cannot be computed directly using finite samples, since $\pi(z)$ is a continuous
random variable. This motivates the need of designing reliable ECE estimators. A popular estimation approach is based on histograms \cite{naeini2015obtaining}. It  partitions the evaluation data points into $b$ bins $\{B_1,\ldots,B_b\}$ according to the predictions $z$, calculate the average prediction $\bar{f}(B_i)$ and label $\bar{\pi}(B_i)$ inside the bins $B_i$, and estimate ECE by: 
\begin{equation}
\label{hist}
\overline{\text{ECE}}^d(f) = \sum_{i=1}^{b}\frac{\#B_i}{n_e}\lVert \bar{f}(B_i)-\bar{\pi}(B_i)\rVert_d^d.
\end{equation}
where $\#B_i$ denotes the number of instances in $B_i$.

Despite its simplicity, histogram-based estimator suffers from several issues. First, it has bias-variance dilemma with respect to the selection of bin amount and edge locations. For example, too few bins lead to under-estimation of ECE \cite{kumar2019verified}, while too many bins leads to noisy estimates as each bin becomes sparsely populated \cite{ashukha2019pitfalls,ding2020revisiting, nixon2019measuring, vaicenavicius2019evaluating}. Therefore, histogram-based ECE estimators are unreliable (e.g., asymptotically biased or noisy) due to their sensitivity to the binning scheme. Unfortunately, a consistently reliable binning selection scheme does not exist \cite{scott1992multivariate,simonoff1997measuring}. Finally, histogram-based estimator is known to converge slower than other advanced non-parametric density estimators \cite{scott1992multivariate}, leading to a data-inefficient estimation of ECE. 

Next (in Sec.~\ref{beyondts}), we discuss the proposed {\em Mix-n-Match} calibration strategies which satisfy the above discussed desiderata. Later (in Sec.~\ref{Eval}), we will address the issue of designing a reliable and data-efficient ECE estimator.

\section{Designing Calibration Methods}

We first present (in Sec.~\ref{apcm}) a general strategy to design provably accuracy-preserving calibration methods. 
Next, we discuss strategies for parametric (in Sec.~\ref{tss}) and non-parametric calibration methods (in Sec.~\ref{ir}) to fulfill remaining desiderata.

\subsection{Accuracy-preserving Calibration}
\label{apcm}
We present a general form of accuracy-preserving calibration maps and validate its accuracy-preserving property.

\begin{definition}[Accuracy-Preserving Calibration Map] Let $g:[0,1]\rightarrow\mathbb{R}_{\geq 0}$ be a non-negative strictly isotonic function. Then, an {\em accuracy-preserving calibration map} is given by: 
\begin{align}
\label{sharet}
T(z) ={(g(z_1),g(z_2),\ldots,g(z_L))}/{ \sum_{l=1}^{L}g(z_l)}.
\end{align}
\label{apc}
\end{definition}
\vspace{-0.2in}
In Eq.~\eqref{sharet}, we apply the same function $g$ to transform all entries in the prediction probability vector $z$ to an un-normalized vector $g(z)=(g(z_1),\ldots,g(z_L))$; and normalize $g(z)$ to a probability simplex $\Delta^L$. The single strictly isotonic function $g$ maintains the ordering of class prediction probabilities, and preserves the classification accuracy. 

\begin{proposition}
\label{ppopta}
The calibration map in Eq.~\eqref{sharet} preserves the classification accuracy of the uncalibrated classifier.
\end{proposition}
\begin{proof}
Please see supplementary material Sec.~\ref{potpa}.
\end{proof}

\subsection{Parametric Calibrations}
\label{tss}
Parametric methods are already data-efficient, thus, one simply needs to enforce the accuracy-preserving requirement and improve their insufficient expressive power.

\subsubsection{Preserving accuracy} As discussed in Proposition~\ref{ppopta}, the use of a strictly isotonic function preserves the accuracy. Fortunately, several existing parametric methods, such as, Platt \cite{platt2000probabilistic} or temperature scaling~\cite{guo2017calibration}, and beta scaling \cite{kull2017beyond}, employ strictly isotonic functions -- logistic function and beta function, respectively. Therefore, these methods are already accuracy-preserving. Otherwise, the general form as provided in Eq.~\eqref{sharet} can be used for designing accuracy-preserving calibration maps.  

\subsubsection{Improving expressivity by model ensemble} 
\begin{figure}[!t]
\centering
\centerline{\includegraphics[width=\columnwidth]{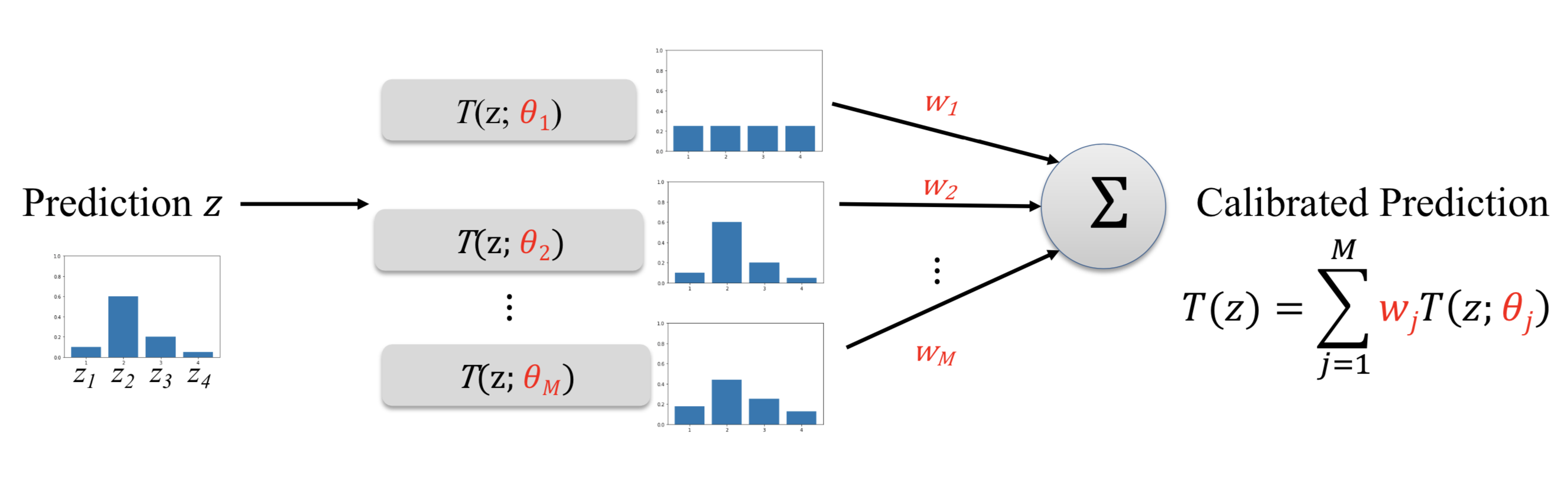}}
\vspace{-0.1in}
\caption{Model Ensemble calibration for improving expressivity. The idea is to use the weighted averaged outputs of an ensemble of $M$ calibration maps to get the final calibrated prediction. Trainable parameters are highlighted in red.}
\label{ms}
\vspace{-0.15in}
\end{figure}

We outline a strategy compatible with any parametric calibration method to improve its expressivity. The idea is to use an ensemble of calibration maps from the same accuracy-preserving parametric family, but with different parameters (see \autoref{ms}): 
$$T(z) = w_1 T(z;\theta_1) +w_2 T(z;\theta_2) + \ldots + w_M T(z;\theta_M),$$ 
where $w$ are non-negative coefficients summing up to one. The weighted sum preserves isotonicity, thus the ensemble inherits the accuracy-preserving property from its individual components. The increased expressivity stems from the fact that more parameters becomes adjustable, including $\theta_j$ and the weights $w_j$ for each component $j$ in the ensemble. We find the weights $w$ and parameters $\theta$ by minimizing the loss $R(.)$ between calibrated predictions $T(z)$ and labels $y$:
\begin{equation*}
\begin{aligned}
& \underset{w,\theta}{\text{minimize}}
& & \sum_{i=1}^{n_c} R\big(\sum_{j=1}^{M}w_j T(z^{(i)};\theta_j),y^{(i)}\big) \\
& \text{s.t.}
& & \mathbf{1}_{1\times M}w=1; w \geq \mathbf{0}_{M\times1}.
\end{aligned}
\end{equation*}

Using the above formulation, we show a specific generalization of temperature scaling (\textbf{TS}) \cite{guo2017calibration} to satisfy the proposed desiderata. 

\textbf{Ensemble Temperature Scaling (ETS).} Note that TS is already accuracy-preserving and data-efficient. Next, we propose an ensemble formulation to improve the expressivity of TS while maintaining its accuracy-preserving and data-efficiency properties. Specifically, we propose a three-component ensemble as follows:
\begin{equation}
T(z;w,t) = w_1 T(z;t) +w_2 z + w_3\frac{1}{L},
\end{equation}
where the calibration map for original TS is expressed by $T(z;t)={(z_1^{1/t},z_2^{1/t},\ldots,z_L^{1/t})}/{\sum_{l=1}^L z_l^{1/t}}$. Interestingly, the remaining two components in the ensemble are also TS calibration maps but with fixed temperature $t$:
\begin{itemize}
    \item TS calibration map with $t = 1$ (outputs {uncalibrated prediction} $z$). It increases the stability when the original classifier is well calibrated \cite{kull2017beyond}.
    \item TS calibration map with $t=\infty$ (outputs {uniform prediction} $z_l=1/L$ for each class). It `smooths' the predictions, similar to how {\em label-smoothing} training technique smooths the one-hot labels \cite{szegedy2016rethinking}, which has shown to be successful in training better calibrated neural networks \cite{muller2019does}.
\end{itemize}

The weight $w$ and temperature $t$ of ensemble is identified by solving the following convex optimization problem:
\begin{equation*}
\begin{small}
\begin{aligned}
& \underset{t,w}{\text{minimize}}
& & \sum_{i=1}^{n_c} R\big(w_1T(z^{(i)};t)+w_2z^{(i)}+w_3\frac{1}{L},y^{(i)}\big) \\
& \text{s.t.}
& & t>0; \mathbf{1}_{1\times3}w=1; w \geq \mathbf{0}_{3\times1}.
\end{aligned}
\end{small}
\end{equation*}

ETS preserves the accuracy, as it uses a convex combination of (strictly) isotonic function $g=z_l^{1/t}$ across all classes/components. Further, as ETS only has three additional parameters (the weights) compared to TS, we expect it to be data-efficient. We will see later in Sec.~\ref{sec:cee}, ETS is significantly more expressive than TS while maintaining its accuracy-preserving and data-efficiency properties.

\subsection{Non-parametric Calibrations}
\label{ir}
Since non-parametric methods are generally expressive, we focus on providing solutions to enforce the accuracy-preserving requirement, and to improve their data-efficiency.

\subsubsection{Preserving Accuracy}
Following Proposition~\ref{ppopta}, in order to preserve accuracy, a strictly isotonic calibration function is needed to be constructed non-parametrically.
For binary classification, this requirement is satisfied by the isotonic regression (IR) calibration \cite{zadrozny2002transforming}: for class $1$ (class $2$ is the complement), it sorts data points according to their predictions ($z_1^{(1)}\leq z_1^{(2)}\ldots\leq z_1^{(n_c)}$), then fits an isotonic function $g$ to minimize the residual between $g(z_1)$ and $y_1$. The common way to extend this method to a multi-class setting is to decompose the problem as $L$ one-versus-all problem, which we further denote as \textbf{IROvA}. Unfortunately, this formulation is neither accuracy-preserving nor data-efficient. 
To extend IR to multi-class problems while preserving accuracy, we use the accuracy-preserving calibration map as defined in Def.~\ref{apc}. This calibration map work identically on all the classes and does not distinguish among them. Next, we explain how this procedure is also more data-efficient than the conventional IROvA approach. 

\begin{figure}[!t]
\centering
\centerline{\includegraphics[width=\columnwidth]{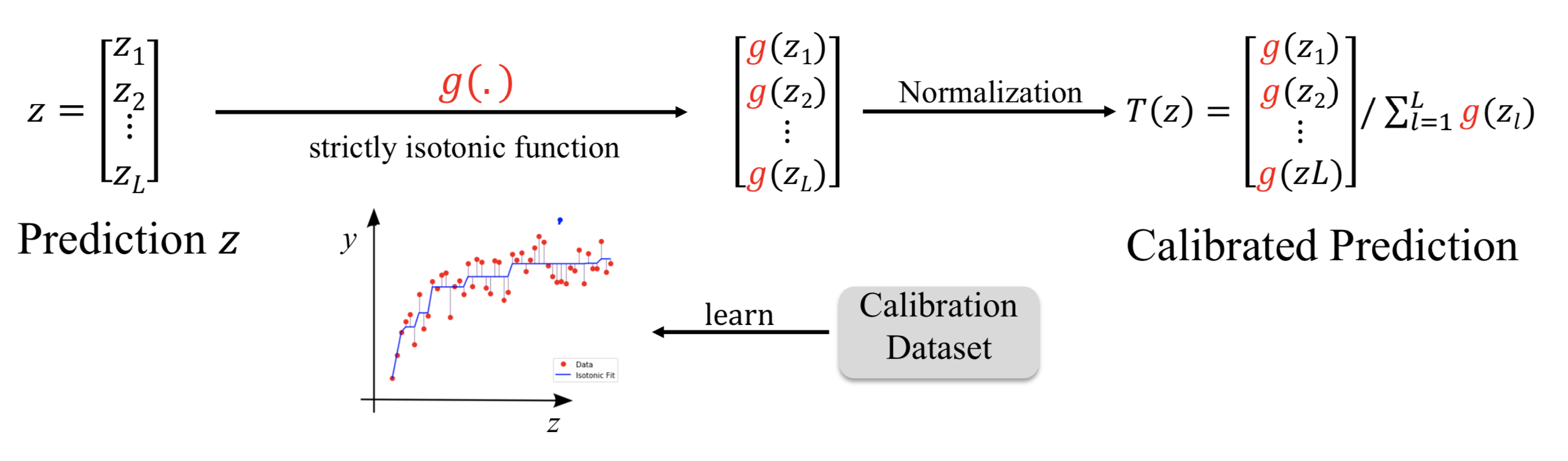}}
\vspace{-0.1in}
\caption{Data Ensemble calibration for improving data-efficiency. The idea is to ensemble the prediction-label pairs from all $L$ classes and learn a single calibration map (e.g., strictly isotonic function for IRM) that is highlighted in red.}
\label{sd-irm}
\vspace{-0.15in}
\end{figure}

\subsubsection{Improving Efficiency by Data Ensemble}
We first explain the proposed multi-class isotonic regression (\textbf{IRM}) procedure, and then comment on its data-efficiency.

IRM first ensembles the predictions and labels from all the classes, then learn a strictly isotonic function $g$ that best fits the transformed predictions versus labels (see \autoref{sd-irm}):

\textbf{Step 1 (Data ensemble)}: extract all entries of prediction vector $\{z^{(i)}\}_{i=1}^{n_c}$ and label vector $\{y^{(i)}\}_{i=1}^{n_c}$. Let $\{a^{(j)}\}_{j=1}^{n_cL}$ and $\{b^{(j)}\}_{j=1}^{n_cL}$ denote the set of $n_cL$ prediction and label entries. Sort both vectors such that $a^{(1)}\leq a^{(2)}\leq \ldots a^{(n_cL)}$.   

\textbf{Step 2 (Isotonic regression)}: learn an isotonic function $g^*$ by minimizing the squared error loss between $g(a)$ and $b$: 
\begin{equation*}
\begin{aligned}
& \underset{g\in \mathcal{G}}{\text{minimize}}
& & \sum_{j=1}^{n_cL} [g(a^{(j)})-b^{(j)}]^2,
\end{aligned}
\end{equation*}
where $\mathcal{G}$ is a family of piecewise constant isotonic functions \cite{zadrozny2002transforming}. The {\em pair-adjacent violator} algorithm \cite{ayer1955empirical} is used to find the best function.

\textbf{Step 3 (Imposing strict isotonicity)}: the learned function $g^*$ is only isotonic. To make it strictly isotonic, we modify it to $\hat{g}(a) = g^*(a) + \epsilon a$, where $\epsilon$ is a very small positive number, such that $g(a)<g(a')$ whenever $a<a'$. Plugging the strictly isotonic function $\hat{g}$ back to Eq.~\eqref{sharet}, we can obtain the non-parametric calibration map.

\textbf{Remark}. Comparing to IROvA, the proposed IRM {preserves the accuracy}. In addition, it is {more data-efficient}, since it uses $n_cL$ data points to identify one isotonic function in contrast to $n_c$ data points in IROvA. 
We should also highlight that these benefits do not come free: by enforcing the same calibration map on all the classes, the proposed approach is {less expressive} than IROvA. In fact, we expect an {\em efficiency-expressivity trade-off} for the proposed solution -- with the number of classes $L$ increasing, it will become more data-efficient but less expressive comparing to one-vs-all. This phenomenon is later verified in Sec.~\ref{sec:cnnc}.

\begin{figure}
\centering
\centerline{\includegraphics[width=\columnwidth]{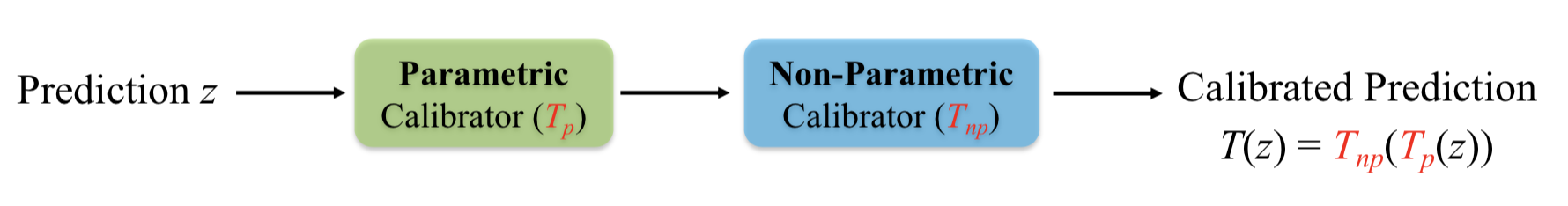}}
\vspace{-0.1in}
\caption{Composition-based calibration for achieving both high data-efficiency as well as high expressivity. The uncalibrated prediction is transformed by the parametric (or efficient) calibrator, followed by the non-parametric (or expressive) calibrator. Trainable parameters are highlighted in red.}
\label{sd-comp}
\vspace{-0.15in}
\end{figure}

\subsection{The Best of Both Worlds by Composition}
Parametric and non-parametric approaches each have their own advantages.
To get the best of both worlds, i.e., high data-efficiency of parametric methods and high expressivity of non-parametric methods, we propose a {\em compositional} method as well. 
Specifically, we propose to apply a data-efficient parametric calibration method first, and then conduct non-parametric calibration on the parametric calibrated entries (see \autoref{sd-comp}). 
Intuitively, first fitting a parametric function acts like a baseline for variance reduction \cite{kumar2019verified} and then conducting a non-parametric calibration enjoys higher data-efficiency than the non-parametric calibration alone. 
Expressivity is unaffected by the composition since no additional restriction is imposed on the non-parametric layer. 
Accuracy-preserving property is satisfied if the adopted parametric and non-parametric calibration maps satisfy Def.~\ref{apc}, since the composition of strictly isotonic functions remains strictly isotonic.

\section{Evaluating Calibration Errors}

\label{Eval}

Next step in the calibration pipeline is to evaluate the calibration performance by estimating the expected calibration error as given in Eq.~\eqref{cali}. The primary challenge is the involvement of two unknown densities $p(z)$ and $\pi(z)$. Histogram-based estimator \cite{naeini2015obtaining} replaces the unknown densities by their bin-discretized version as given in Eq.~\eqref{hist}. It is easy to implement, but also inevitably inherits drawbacks from histograms, such as the sensitivity to the binning schemes, and the data-inefficiency.

We alleviate these challenges by replacing histograms with non-parametric density estimators that are continuous (thus, avoid the binning step) and, are more data-efficient.
Specifically, we use kernel density estimation (KDE) \cite{parzen1962estimation, rosenblatt1956remarks} to estimate the ECE for its implementation easiness and tractable theoretical properties.

\subsection{KDE-based ECE Estimator}
Let $K: \mathbb{R} \rightarrow \mathbb{R}_{\geq 0}$ denote a smoothing {\em kernel function} \cite{tsybakov2008introduction}. Given a fixed bandwidth $h>0$, we have $K_h(a)=h^{-1}K(a/h)$. Based on the evaluation dataset, the unknown probabilities are estimated using KDE as follows:
\vspace{-.2in}
\begin{align*}
\begin{split}
\tilde{p}(z) &= \dfrac{h^{-L}}{n_e}\sum\limits_{i=1}^{n_e} \prod_{l=1}^{L}K_h(z_l-z_l^{(i)}),\\
\tilde{\pi}(z) &= \dfrac{\sum\limits_{i=1}^{n_e} y^{(i)} \prod_{l=1}^{L}K_h(z_l-z_l^{(i)})}{\sum\limits_{i=1}^{n_e} \prod_{l=1}^{L}K_h(z_l-z_l^{(i)})}.
\end{split}
\label{kde1}
\end{align*}
Plugging them back in Eq.~\eqref{cali}, we obtain the KDE-based ECE estimator:
\begin{equation}
\label{np}
\widetilde{\text{ECE}}^d(f) = \int \lVert z-\tilde{\pi}(z)\rVert_d^d\;\tilde{p}(z)\mathop{dz}.
\end{equation}
The integration in Eq.~\eqref{np} can be performed numerically (e.g., using Trapzoidal rule).

We next provide a theoretical analysis of statistical properties of the proposed KDE ECE estimator when $d=1$. The results for $d=2$ can be obtained similarly.

\begin{theorem}[Statistical properties] Assuming the unknown densities $p(z)$ and $\pi(z)$ are smooth ($\beta$-H\"older) and bounded, with the bandwidth $h\asymp n_e^{-1/(\beta+L)}$, the KDE ECE is asymptotically unbiased and consistent, with a convergence rate $|\mathbb{E}[\widetilde{\text{ECE}}^1(f)]-\text{ECE}^1(f)| \in O(n_e^{-\beta/(\beta+L)})$.
\label{pythagorean}
\end{theorem}
\vspace{-.2in}
\begin{proof}
Please see supplementary material Sec.~\ref{sec:kdeproof}.
\end{proof}


As verifying these smoothness assumptions in practice is highly non-trivial~\cite{kumar2019verified}, we corroborate our theoretical results using empirical comparisons in Sec.~\ref{sec:cee}. 
The implementation details for KDE is provided in the supplementary material Sec.~\ref{kid}.

\textbf{Dimensional reduction for multi-class problems}. Convergence rates of non-parametric density estimators depend undesirably on the class dimension $L$, making the estimation challenging for multi-class problems. A way around this curse of dimensionality problem is to use the \emph{top-label} ECE$^d$ \cite{guo2017calibration}
or the {\em class-wise} ECE$^d$ \cite{kull2019beyond,kumar2019verified}.
Both reduce the effective dimension to one, but weaken the calibration notion in Def.~\ref{perfect}, meaning that they can be zero even if the model is not perfectly calibrated \cite{vaicenavicius2019evaluating}.

\subsection{A Dimensionality-Independent Ranking Method}
In many practical situations, the main goal for evaluating calibration errors is to compare (or rank) calibration maps. However, rankings based on the approximations, e.g., top-label and class-wise ECE$^d$, have been observed to be contradictory \cite{kull2019beyond,nixon2019measuring}. This raises the question rankings based on these approximations are indicative of the ranking based on actual ECE$^d$ in Eq.~\eqref{cali}. 

Next, we present a dimensionality-independent solution to compare calibration maps according to their actual calibration capabilities, rather than resorting to weaker variants. The solution relies on the well-known {\em calibration refinement decomposition} \cite{murphy1973new} for the {\em strictly proper scoring loss} \cite{gneiting2007strictly}. Thus, it is applicable only when $d=2$, since the absolute loss ($d=1$) is improper \cite{buja2005}. Since ECE$^1$ and ECE$^2$ are closely related ($\sqrt{\text{ECE}^2}<$ECE$^{1}<\sqrt{L\cdot\text{ECE}^2}$), we anticipate comparisons based on ECE$^2$ and ECE$^1$ should be similar. Specifically, we propose to use calibration gain (defined next) for the comparison.
\begin{definition}
The {\em calibration gain} is defined as the reduction in ECE$^d$ after applying a calibration map ($T\circ f$):
\begin{equation*}
\Delta\text{ECE}^2(T) = \text{ECE}^2(f)-\text{ECE}^2(T\circ f).
\end{equation*}
\end{definition}
Higher gain indicates a better calibration map. 
\begin{proposition}
\label{propapc}
For accuracy-preserving maps in Def.~\ref{apc}, the calibration gain equals the reduction of squared loss between predictions and labels after calibration:
\begin{equation}
\label{ecet}
\Delta\text{ECE}^2(T) = \mathbb{E}\lVert Z-Y\rVert_2^2-\mathbb{E}\lVert T(Z)-Y\rVert_2^2.
\end{equation}
\end{proposition}
\begin{proof}
Please see supplementary material Sec.~\ref{potceg}.
\end{proof}

For non accuracy-preserving methods (\autoref{top_cali}), the squared loss reduction in Eq.~\eqref{ecet} bounds its actual calibration gain from below, and may not facilitate a fair comparison.

Finally, given an evaluation dataset, Eq.~\eqref{ecet} is estimated by:
\begin{equation}
\label{cg}
\Delta \widehat{\text{ECE}}^2 (T) = \frac{1}{n_e}{\sum\limits_{i=1}^{n_e}\big(\lVert z^{(i)}-y^{(i)}\rVert_2^2-\lVert T(z^{(i)})-y^{(i)}\rVert_2^2\big)}   
\end{equation}
which converges at the rate of $O(n_e^{-1/2})$ independent of the class dimension $L$ and avoids the curse of dimensionality.

\section{Experiments}

\subsection{Calibration Error Evaluations}
\label{sec:cee}

\begin{table*}[!t]
\caption{Top-label ECE$^1$ (\%) (lower is better). The number following a model’s name denotes the network depth (and width if applicable).}
\label{top_cali}
\vskip 0.15in
\begin{center}
\begin{small}
\begin{tabular}{llcccccccc}
\toprule
Dataset & Model & Uncalibrated & TS & ETS & IRM & IROvA & IROvA-TS & DirODIR & GPC \\
 &  &  &  & (\textit{ours}) & (\textit{ours}) &  &  (\textit{ours}) & & \\
\midrule
CIFAR-10 & DenseNet 40 & 3.30 & \textbf{1.04} & \textbf{1.04}  & 1.18 & 1.16  & 1.11 & 1.23 & 1.68 \\
CIFAR-10 & LeNet 5 & 1.42 & 1.16  & \textbf{1.13} & 1.19 & 1.26 & 1.26 & 1.29 & 1.14 \\
CIFAR-10 & ResNet 110 & 4.25 & 2.05  & 2.05  & 1.53  & 1.45  & \textbf{1.39} & 1.82 & 1.40 \\
CIFAR-10 & WRN 28-10 & 2.53  & 1.61 & 1.61 & 1.02  & 0.994 & \textbf{0.967} & 1.49 & 1.05 \\
\midrule
CIFAR-100 & DenseNet 40 & 12.22  & 1.55  & 1.54  & 3.32  & 4.48 & 2.22 & 1.56 & \textbf{1.51} \\
CIFAR-100 & LeNet 5 & 2.76  & 1.11  & \textbf{1.05} & 1.33  & 3.67 & 3.18 & 1.17 & 1.36 \\
CIFAR-100 & ResNet 110 & 13.61  & 2.75  & \textbf{1.93} & 4.78 & 5.27 & 3.00 & 2.46 & 1.98 \\
CIFAR-100 & WRN 28-10 & 4.41 & 3.24  & 2.80 & 3.16  & 3.45 & 2.92 & 3.11 & \textbf{1.58} \\
\midrule
ImageNet & DenseNet 161 & 5.09 & 1.72 & \textbf{1.33} & 2.13  & 3.97 & 3.01 & 4.61 & - \\
ImageNet & ResNeXt 101 & 7.44 & 3.03 & \textbf{2.02} & 3.51  & 4.64 & 3.09 & 5.02 & - \\
ImageNet & VGG 19 & 3.31 & 1.64 & \textbf{1.36} & 1.85 & 3.77 & 3.03 & 4.04 & - \\
ImageNet & WRN 50-2 & 4.83 &  2.52  & \textbf{1.81} & 2.54 & 3.91  & 3.03 & 4.80 & -  \\
\bottomrule
\end{tabular}
\end{small}
\end{center}
\vskip -0.1in
\end{table*}

\begin{table*}[!t]
\caption{Calibration Gain $\Delta$ECE$^2$ (\%) (higher is better). Reported $\Delta$ECE$^2$ underestimate the actual calibration gains for IROvA, IROvA-TS, DirODIR, GPC. The number following a model’s name denotes the network depth (and width if applicable).}
\label{top_cali1}
\vskip 0.15in
\begin{center}
\begin{small}
\begin{tabular}{llccccccc}
\toprule
Dataset & Model & TS & ETS & IRM & IROvA & IROvA-TS & DirODIR & GPC \\
 &  &  &  (\textit{ours}) & (\textit{ours}) &  &  (\textit{ours}) & & \\
\midrule
CIFAR-10 & DenseNet 40 & \textbf{0.611} & \textbf{0.611} & 0.562  &  0.560  & 0.593  & 0.606 & 0.457 \\
CIFAR-10 & LeNet 5 & 0.027 & \textbf{0.028}  & -0.028 & -0.004  & -0.007   & -0.119 & 0.022  \\
CIFAR-10 & ResNet 110 & 0.821 & 0.821 & 0.976  & 1.11  & \textbf{1.15}  & 1.13   & 1.02 \\
CIFAR-10 & WRN 28-10 & 0.403  & 0.403 & 0.596 & 0.614  & 0.617   & 0.297 & \textbf{0.624}\\
\midrule
CIFAR-100 & DenseNet 40 & 2.74  & \textbf{2.75}  & 2.50  & 1.99  & 2.17    & 2.36 & 2.57 \\
CIFAR-100 & LeNet 5 & 0.077  & \textbf{0.085}  & 0.028 & -0.576  & -0.558  & -0.445 & 0.055  \\
CIFAR-100 & ResNet 110 & 3.14 & 3.17  & 2.90  & 2.63  & 3.09 &  \textbf{3.50} & 3.14  \\
CIFAR-100 & WRN 28-10 & 0.204 & 0.263  & 0.534 & 0.134  & 0.218 & 0.0289 & \textbf{0.841} \\
\midrule
ImageNet & DenseNet 161 & 0.397 & \textbf{0.423} & 0.368 & -0.518  & -0.438 & -1.63 & - \\
ImageNet & ResNeXt 101 & 0.915 & \textbf{0.995} & 0.90 & 0.028  & 0.233 & -1.35 & - \\
ImageNet & VGG 19 & 0.147 & \textbf{0.168} & 0.115 & -0.989 & -0.969 &  -1.68 & - \\
ImageNet & WRN 50-2 & 0.266 &  0.317  & \textbf{0.321} & -0.604 & -0.543  & -1.51 & -\\
\bottomrule
\end{tabular}
\end{small}
\end{center}
\vskip -0.1in
\end{table*}

\begin{figure}[!t]
\centering
\centerline{\includegraphics[width=\columnwidth]{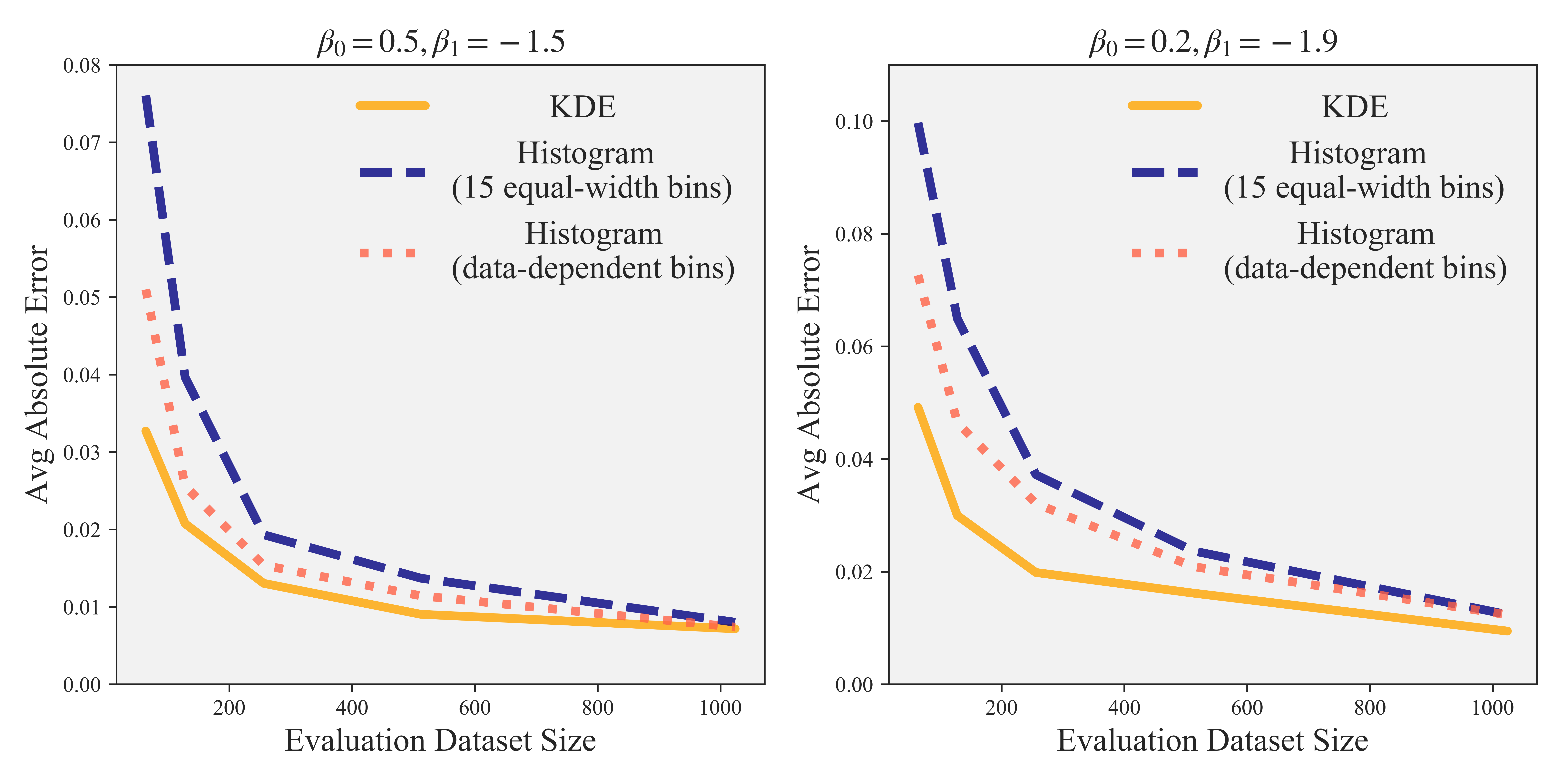}}
\vspace{-0.1in}
\caption{Average absolute error for different ECE estimators as a function of the number of samples used in the ECE estimation. KDE-based ECE estimator achieves lower estimation error, especially when the evaluation dataset is small.}
\label{f61}
\vspace{-0.15in}
\end{figure}

We compare the finite sample performance of proposed KDE-based ECE$^d$ estimator with histogram-based ones on a synthetic binary classification problem ~\cite{vaicenavicius2019evaluating}. The classifier is parameterized by two parameters $\beta_0,\beta_1$ (described in detail in supplementary material Sec.~\ref{sm}). We consider a less-calibrated case $\beta_0=0.5,\beta_1=-1.5$, and a better-calibrated case with $\beta_0=0.2,\beta_1=-1.9$. The canonical calibration probability $\pi^{(f)}(z)$ has a closed-form expression (Eq.~\eqref{mixcali}), allowing us to obtain the ground truth ECE (Eq.~\eqref{cali}) using Monte Carlo integration with $10^6$ samples. We compare the ground truth to the estimation of ECE$^1$ using KDE and histogram-based estimators -- one with $15$ equal-width bins and other with data-dependent binning scheme \cite{sturges1926choice}. In \autoref{f61}, we vary the size of evaluation samples $n_e$ from $64$ to $1024$ and plot the mean absolute error (averaged over $1000$ independent experiments) between KDE/Histograms estimates and the ground truth. KDE consistently outperforms histogram-based estimators regardless of the binning schemes. The discrepancy is particularly noticeable with small $n_e$, highlighting KDE's superior efficiency in data-limited regime. In rest of the paper, we adopt KDE for estimating ECE, unless otherwise specified. 
Additional results can be found in the supplementary material, e.g., the distribution of the estimation errors in \autoref{f62} {and comparison of the proposed KDE estimators with recently proposed debiased histogram ECE estimators \cite{kumar2019verified} in \autoref{debias}}. 

\subsection{Calibrating Neural Network Classifiers}
\label{sec:cnnc}
We calibrate various deep neural network classifiers on popular computer vision datasets: CIFAR-10/100 \cite{krizhevsky2009learning} with 10/100 classes and ImageNet \cite{deng2009imagenet} with 1000 classes. For CIFAR-10/100, we trained DenseNet \cite{huang2017densely}, LeNet \cite{lecun1998gradient}, ResNet \cite{he2016deep} and WideResNet (WRN) \cite{zagoruyko2016wide}. The training detail is described in Sec.~\ref{sec:ar}. We use 45000 images for training and hold out 15000 images for calibration and evaluation. For ImageNet, we acquired 4 pretrained models from \cite{paszke2019pytorch} which were trained with $1.3$ million images, and $50000$ images are hold out for calibration and evaluation.

We compare seven calibration methods: for parametric approaches, we use TS, our proposed three-component model ensemble approach ETS, and the Dirichlet calibration with off-diagonal regularization (DirODIR) \cite{kull2019beyond}. Following \cite{kumar2019verified}, we use the squared error as the loss function to fit TS and ETS. For non-parametric approaches, we compare IROvA, our proposed multi-class accuracy-preserving scheme IRM, and the composition method that combines IROvA with TS as described in Sec.~\ref{ir} (referred to as \textbf{IROvA-TS}). In addition, we include the Gaussian Process calibration (GPC) \cite{wenger2019non}. Among all examined methods, only TS, ETS and IRM are accuracy-preserving.

\begin{figure*}[!t]
\vskip 0.2in
\centering
\centerline{\includegraphics[width=1.65\columnwidth]{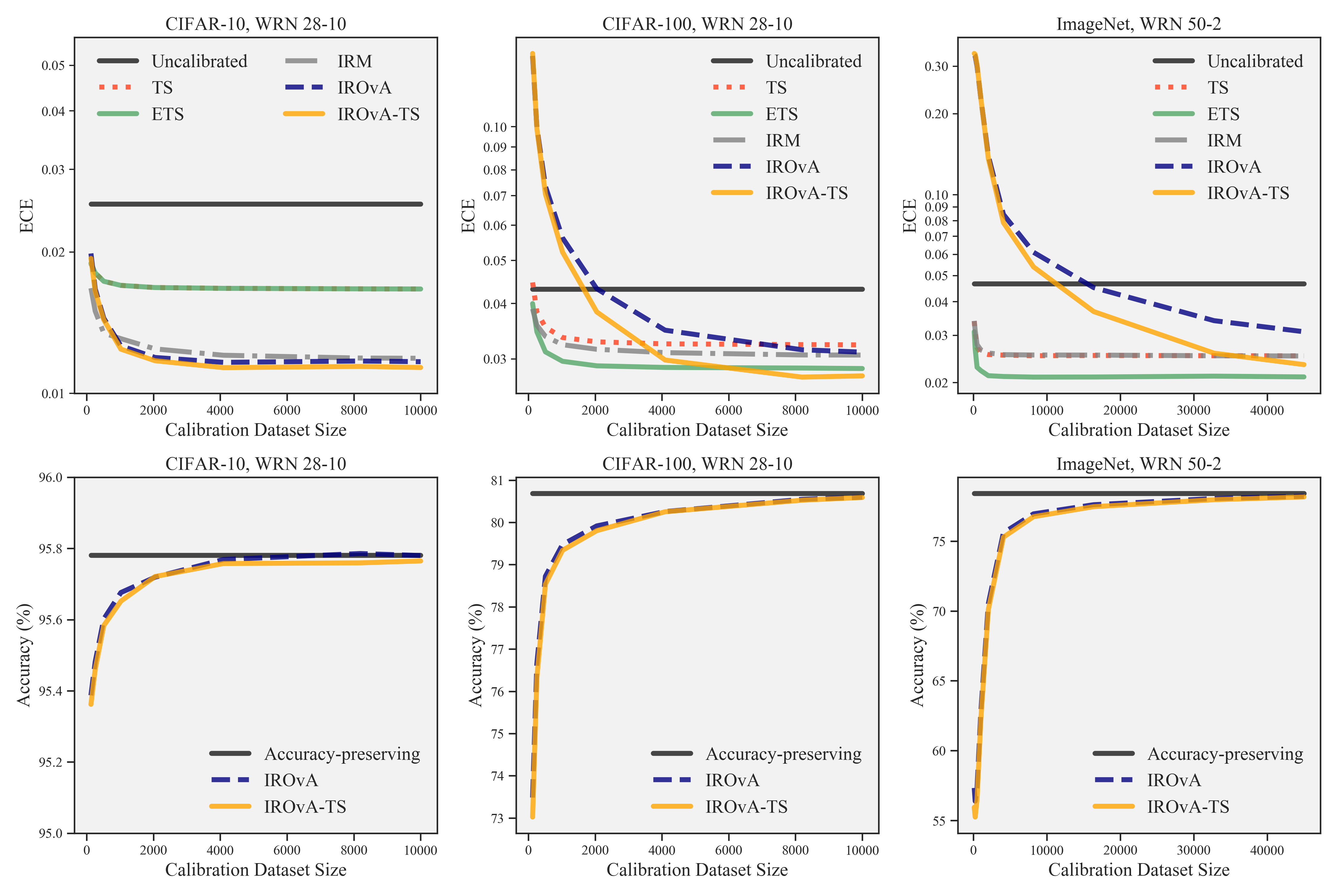}}
\caption{Learning curve comparisons of various calibration methods on top-label ECE$^1$ (top) and the classification accuracy (bottom).}
\label{lcimagenet}
\vskip -0.2in
\end{figure*}

For our first experiment, we adopt a standard calibration setup \cite{guo2017calibration} with fixed-size calibration $n_c$ and evaluation $n_e$ datasets. We randomly split the hold-out dataset into $n_c=5000$, $n_e= 10000$ for CIFAR-10/100 and $n_c=n_e=25000$ for ImageNet. We randomly split the hold-out dataset into $n_c$ calibration points to learn the calibration map, and $n_e$ evaluation points to evaluate ECE and classification accuracy. All results are averaged over $100$ independent runs. On ImageNet, GPC fails to converge due to its high computational cost, thus we exclude its results.

\autoref{top_cali} displays top-label ECE$^1$ and \autoref{top_cali1} displays the calibration gain $\Delta$ECE$^2$. Overall the rankings of calibration methods by top-label  ECE$^1$ or by $\Delta$ECE$^2$ are very similar. Our proposed strategies consistently lead to better performance than the baseline implementations (ETS over TS; IRM and IROvA-TS over IROvA). 
{Depending on the model/data complexity, either parametric or non-parametric variants may be more suitable.}

{Note that calibration methods may perform differently with varying amounts of calibration data. A critically missing aspect of the standard practice of fixed-size comparison is that it does not reveal the data-amount-dependent behavior, and it may provide an incomplete picture of the calibration method's performance. To explore this holistically, we next conduct a learning curve analysis by varying the calibration dataset size and evaluating three desiderata-related properties (accuracy, data-efficiency and expressivity) of calibration approaches. While such learning curve analysis has been extensively used in the standard machine learning literature, its use in the calibration of deep neural network classifiers is scarce. Specifically, we reserve the same set of $5000$ data points for evaluation, and vary the number of calibration data from $128$ to $10000$ (CIFAR-10/100) or $45000$ (ImageNet). This process is repeated 100 times on the baseline (TS, IROvA) and their variants (ETS, IRM, IROvA-TS) to validate the effectiveness of {\em Mix-n-Match}.}

For Wide ResNets, \autoref{lcimagenet} shows how the average ECE and the accuracy over the repetitions change as a function of calibration dataset size $n_c$. Results for other cases are provided in the supplementary material Sec.~\ref{sec:ar}. 
We provide a thorough analysis on the learning curves below.

\textbf{Accuracy}. From \autoref{lcimagenet} bottom, we observe that IROvA and IROvA-TS lead to serious accuracy reduction in the data-limited regime, and require a large calibration datasets to recover the original accuracy, while the accuracy-preserving approaches maintain the accuracy.

\textbf{Data-efficiency}. Parametric methods (TS, ETS) enjoy the fastest converge of ECE (see \autoref{lcimagenet} top), which is anticipated. Our proposed data ensemble (IRM) and compositional (IROvA-TS) solutions also converge faster than IROvA. To quantify their data-efficiency gain, we record the required amount of data for non-parametric approaches to reach a reference calibration level (see \autoref{quant-measure} right) in \autoref{efficient}. The proposed data ensemble and compositional approaches achieve remarkable data-efficiency gains. 

\textbf{Expressivity}. Note that a more expressive method should attain lower ECE with a sufficiently large calibration dataset.
From \autoref{lcimagenet} top, we see that the proposed ensemble approach ETS is significantly more expressive than TS. This gain is particularly visible in many-class datasets, e.g.,  CIFAR-100 and ImageNet, where the canonical calibration function is expected to be complex. Also, the reduced expressivity of IRM over IROvA can be observed, verifying our hypothesis of its efficiency-expressivity trade-off. In \autoref{expressive}, we provide a quantitative comparison on the expressivity of TS and ETS by measuring their final ECE (at the highest values of $n_c$, see \autoref{quant-measure} left). We see that ETS is consistently more expressive and achieves lower final ECE value than TS across different models/datasets.

{Finally, we provide general guidelines on choosing an appropriate calibration method. We recommend {ETS} for general use and {IRM} as a strong alternative when the parametric form of ETS is misspecified (see \autoref{lcimagenet} left). Both approaches are accuracy-preserving and can be compared based on the proposed {calibration gain} metric. For complex calibration tasks, we recommend {IROvA-TS} if a large calibration dataset is available and the user does not have hard constraints on preserving the accuracy (see \autoref{lcimagenet} middle-right). 
We expand the analysis and the recommendation in the supplementary material Sec.~\ref{sec:guide}.}



To summarize, our proposed {\em{Mix-n-Match}} strategies provide substantial benefits for calibration, and can be easily incorporated into many existing calibration methods. 
We also provide guidelines on determining the most appropriate calibration method for a given problem in Sec.~\ref{sec:guide}. We expect the observed trends to generalize to other potential extensions. For example, the ensemble beta scaling method should be more expressive than the original beta scaling method \cite{kull2017beyond}, and the composition of TS with other non-parametric methods, e.g., IRM, should also be more data-efficient. We also anticipate additional efficiency gain if one substitutes TS with ETS in the composition, since ETS has been shown to be more expressive.


\FloatBarrier

\section{Conclusion}

We demonstrated the practical importance of designing calibration methods with provable accuracy-preserving characteristics, high data-efficiency, and high expressivity. We proposed general {\em Mix-n-Match} calibration strategies (i.e., ensemble and composition) to extend existing calibration methods to fulfill such desiderata simultaneously. Furthermore, we proposed a data-efficient kernel density-based estimator for a reliable evaluation of the calibration performance. Comparisons with existing calibration methods across various datasets and neural network models showed that our proposed strategies consistently outperform their conventional counterparts. We hope that our developments will advance research on this essential topic further.

\section*{Acknowledgements}
This work was performed under the auspices of the U.S. Department of Energy by Lawrence Livermore National Laboratory under Contract DE-AC52-07NA27344 and LLNL-LDRD Program Project No. 19-SI-001.


\FloatBarrier
\bibliography{ms}
\bibliographystyle{icml2020}

\clearpage

\onecolumn
\begin{appendices}

\section{Proofs of Proposition~\ref{ppopta}: Accuracy-Preserving Calibration Maps}
\label{potpa}
For an arbitrary pair of classes ($\forall i,j\in[L]$, where $[L]$ denotes the set of positive integers up to $L$) of the prediction probability vector z, let us assume that $z_i<z_j$ without loss of generality. By definition, the strictly isotonic function $g$ will output $g(z_i)<g(z_j)$ after the transformation. After dividing by the same normalization constant $G(z)$, the following relationship holds: $[T(z)]_i< [T(z)]_j$, where $[.]_{i}$ represents the $i$-th entry of the vector. Since the order of entries in the prediction vector is unchanged after the calibration, the classification accuracy is preserved.

\section{Proofs of Theorem~\ref{pythagorean}: Statistical Properties of KDE-based ECE}
\label{sec:kdeproof}

\textbf{Mirror image KDE for boundary correction}. Considering that we work with the probability simplex $\Delta^L$ in the context of calibration, KDE can suffer from excessively large bias near the boundary of the simplex. To suitably correct for boundary bias without compromising the estimation quality, we adopt the {\em mirror image} KDE strategy \cite{singh2014generalized}. The convergence/consistency properties will be proved for such a choice. 

\textbf{Smoothness assumption on the underlying densities}. Let $\beta$ and $N$ be positive numbers. Given a vector $s=(s_1,\ldots,s_L)$ with non-negative integer entries, let us use $D^{s}:=\frac{\partial^{\lVert s \rVert_1}}{\partial^{s_1} z_1 \ldots \partial^{s_L} z_L}$ to denote the differential operator. The $\beta$-H\"older class of densities $\Sigma(\beta,N)$ contains those densities $p:[0,1]^L\rightarrow\mathbb{R}$ satisfying the following relationship:
\begin{align*}
|D^sp(z)-D^sp(z')|\leq N \lVert z-z' \rVert^{\beta-\lVert s \rVert_1},
\end{align*}
for all $z,z' \in \mathcal{Z}$ and all $s$ with $\lVert s \rVert_1=\beta-1$. We assume that the distribution for predictions $p(z)$ as well as the calibration probabilities $\pi_l$ for each class $l\in[L]$ belongs to the $\beta$-H\"older class. 

\textbf{Assumption on the kernel function}. We assume that the kernel function $K:\mathbb{R} \rightarrow \mathbb{R}_{\geq 0}$ has bounded support $[-1, 1]$ and satisfies:
\begin{align*}
&\int_{-1}^1K(u)\mathop{du}=1; \lVert K \rVert_1 = \int_{-1}^1|K(u)|\mathop{du}<\infty;\forall j \in [\beta-1], \int_{-1}^1u^jK(u)\mathop{du}=1.
\end{align*}

\textbf{Boundedness assumption}. We denote $C_{\pi}:=\text{sup}_z\lVert z-\pi(z)||_1$ and $C_{z}:=\text{sup}_z\tilde{p}(z)$ and assume they are both finite.

We can bound the KDE estimation error of $|\text{ECE}(f)-\widetilde{\text{ECE}}(f)|$ after applying the triangle inequality:
\begin{equation}
\label{ineq}
\begin{split}
 &|\text{ECE}(f)-\widetilde{\text{ECE}}(f)| = \bigg|\int \lVert z-\pi(z)\rVert_1 p(z) \mathop{dz}-\int \lVert z-\tilde{\pi}(z))\rVert_1 \tilde{p}(z) \mathop{dz}\bigg|\\
&\leq \int \bigg|p(z) \lVert z-\pi(z)\rVert_1 -\tilde{p}(z)\lVert z-\pi(z)\rVert_1 \bigg|\mathop{dz}+\int \bigg|\tilde{p}(z)\big(\lVert z-\pi(z)\rVert_1-\lVert z-\tilde{\pi}(z)\rVert_1\big)\bigg|\mathop{dz}\\
&\leq \text{sup}_z \lVert z-\pi(z) \rVert_1\int | p(z)-\tilde{p}(z)|\mathop{dz}+\text{sup}_z \tilde{p}(z)\int \lVert \pi(z)-\tilde{\pi}(z)\rVert_1\mathop{dz}\\
& \leq C_{\pi}\int | p(z)-\tilde{p}(z)|\mathop{dz}+C_z\int\lVert \pi(z)-\tilde{\pi}(z)\rVert_1\mathop{dz}
\end{split}
\end{equation}

which connects the absolute estimation error of ECE to the integrated estimation errors on the unknown densities $p(z)$ and $\pi(z)$. We then borrow the established convergence rate and consistency proofs for (conditional) density functional of mirror KDE \cite{singh2014exponential} to derive the statistical properties for the proposed KDE-based ECE estimator.

\textbf{Bias convergence rate}. Taking the expectation and applying the Fubini's theorem on both sides in Eq.~\eqref{ineq}, we can derive:
\begin{align*}
&\mathbb{E}|\text{ECE}(f)-\widetilde{\text{ECE}}(f)| \leq C_{\pi} \int \mathbb{E}|p(z)-\tilde{p}(z)|\mathop{dz}+C_z\int \mathbb{E}\lVert \pi(z)-\tilde{\pi}(z))\rVert_1\mathop{dz}\\
& \leq C_{\pi} C_{B1}(h^\beta+h^{2\beta}+\frac{1}{n_eh^L})+C_zC_{B_2}(h^\beta+h^{2\beta}+\frac{1}{n_eh^L}) \leq C(h^\beta+h^{2\beta}+\frac{1}{n_eh^L}),
\end{align*}
where $C_{B1}$ and $C_{B2}$ are constants given the sample size $n_e$ and bandwidth $h$ and $C=C_{\pi}C_{B1}+C_zC_{B2}$. The quantity $h^{2\beta}$ is introduced by the Bias Lemma \cite{singh2014generalized} from the mirror image KDE. For $\int \mathbb{E}|p(z)-\tilde{p}(z)|\mathop{dz}$, we follow the standard KDE results (see Prop 1.1,1.2 and 1.2.3 \cite{tsybakov2008introduction}) while the bound on the other term $\int \mathbb{E}\lVert \pi(z)-\tilde{\pi}(z))\rVert_1\mathop{dz}$ follows 6.2 in \cite{singh2014exponential} or \cite{doring2016exact,gyorfi2006distribution}. 

The optimal bandwidth is $h\asymp n_e^{-1/(\beta+L)}$, leading to a convergence rate of $O\big(n_e^{-\beta/(\beta+L)}\big)$.

\textbf{Consistency}. Let $\tilde{p}'$ denote the KDE marginal density of $z$ when an existing sample point is replaced by a new sample from the same distribution $p(z)$, and similarly $\tilde{\pi}'$ denote the KDE canonical calibration function after replacing a sample. Following \cite{singh2014exponential}, we can bound the density discrepancy before/after replacing a single sample by: 
\begin{align}
\label{perturb}
\int|\tilde{p}(z)-\tilde{p}'(z)| \mathop{dz} \leq \frac{C_{V1}}{n_e};  \int\lVert \tilde{\pi}(z)-\tilde{\pi}'(z)\rVert_1 \mathop{dz} \leq \frac{C_{V2}}{n_e},
\end{align}
where $C_{V1}$ and $C_{V2}$ are constants in the class-dimension $L$ and the kernel norm $\lVert K \rVert_1$ for exact values. 
Suppose that we use two sets of $n_e$ independent samples to estimate $p$ and $\pi$, respectively. Since $\widetilde{\text{ECE}}(f)$ depends on $2n_e$ independent variables, combining Eq.~\eqref{perturb} with Eq.~\eqref{ineq}, we can use McDiarmid’s Inequality \cite{mcdiarmid1989method} to derive that:
\begin{align*}
\mathbb{P}(|\widetilde{\text{ECE}}(f)-\mathbb{E}\widetilde{\text{ECE}}(f)|> \varepsilon) & \leq 2\exp\bigg(-\frac{2\varepsilon^2}{2n_e (2C_V/n_e)^2}\bigg)= 2\exp\bigg(-\frac{\varepsilon^2n_e}{4C_V^2}\bigg).
\end{align*}
for $C_V=\max (C_{V1},C_{V2})$. As $\mathbb{P}(|\widetilde{\text{ECE}}(f)-\mathbb{E}\widetilde{\text{ECE}}(f)|> \varepsilon)$ approaches 0 when $n_e \rightarrow \infty$, the KDE-based ECE estimator is consistent.



\section{KDE Implementation Detail}
\label{kid}
\textbf{Kernel function choice}. Different types of kernel functions $K(u)$ can be used, such as the Gaussian and Epanechnikov functions. Our choice is the Triweight Kernel $K_h(u)=(1/h)\frac{35}{32}(1-(u/h)^2)^3$ on $[-1, 1]$, since it has been recommended for problems with limited support interval \cite{de1999use}. 

\textbf{Bandwidth selection.} We use the popular rule-of-thumb $h=1.06\hat{\sigma}n_e^{-1/5}$ \cite{scott1992multivariate}, where $\hat{\sigma}$ is the standard deviation of the samples.

\section{Proof for Proposition~\ref{propapc}: Calibration Gain for Accuracy-Preserving Methods}
\label{potceg}
According to the {\em calibration refinement decomposition} \cite{murphy1973new}, the expected calibration error ECE$^{2}$ is equal to:
\begin{equation}
\label{decom}
\text{ECE}^2(f)=  \mathbb{E}\lVert z-\pi(z)\rVert_2^2 = \mathbb{E}\lVert z-y\rVert_2^2-\mathbb{E}\lVert \pi(z)-y\rVert_2^2,
\end{equation}
where $\mathbb{E}\lVert z-y\rVert_2^2$ is the standard square loss and $\mathbb{E}\lVert \pi(z)-y\rVert_2^2$ is the {\em refinement error} \cite{murphy1973new} that penalizes the existence of inputs sharing the same prediction but different class labels. Before proceeding further, we first introduce the definition of {\em injective} calibration maps: 
\begin{definition}[Injective Calibration Map]
The calibration map is {\em injective} if different prediction vectors remain different after calibration: $\forall z,z' \in \mathcal{Z}, T(z)\neq T(z) \text{ if } z \neq z'$.
\end{definition}

\begin{proposition}
The accuracy-preserving calibration map $T$ in Def.~3.1is injective.
\end{proposition}
\begin{proof}
Given $z \neq z'$, without loss of generality assume $G(z) \geq G(z')$ for their normalization constants. Since $z \neq z'$, there must exists at least one class $l$ where $z_l<z_l'$. After the transformation by a strictly isotonic function $g$, we know that $g(z_l)<g(z'_l)$. Then we can derive that:
\begin{equation*}
[T(z)]_l-[T(z')]_l=\frac{g(z_l)}{G(z)}-\frac{g(z_l')}{G(z')}=\frac{g(z_l)G(z')-g(z_l')G(z)}{G(z)G(z')}<0.
\end{equation*}
Therefore, $T(z) \neq T(z')$ because their $l$-th entry is not equal. The calibration map is then injective.

Note that the canonical calibration function in Eq.~\eqref{pi} is essentially the conditional expectation of binary random variables $Y$, as $\pi_l(z)=\mathbb{P}[Y_l=1|f(X)=z]=\mathbb{E}[Y_l|f(X)=z]$. By elementary properties of the conditional expectation, one can easily show that injective calibration maps will not change the canonical calibration probabilities, thus $\pi(z) = \pi(T(z))$ for injective $T$. Combining this with the decomposition relationship in Eq.~\eqref{decom}, we can show that:
\begin{align}
\label{decom1}
\begin{split}
\Delta\text{ECE}^2(T)&=\text{ECE}^2(f)-\text{ECE}^2(T\circ f)\\
&= \mathbb{E}\lVert z-y\rVert_2^2-\mathbb{E}\lVert \pi(z)-y\rVert_2^2 - \big(\mathbb{E}\lVert T(z)-y\rVert_2^2-\mathbb{E}\lVert \pi(T(z))-y\rVert_2^2\big)\\
&=\mathbb{E}\lVert z-y\rVert_2^2-\mathbb{E}\lVert T(z)-y\rVert_2^2.
\end{split}
\end{align}
Therefore, after applying an injective calibration map, which include the proposed accuracy-preserving ones, any changes in the squared loss will be due to the change in ECE$^2$.
\end{proof}

\textbf{Remark}. Most existing calibration methods are not injective. For example, in histogram binning \cite{zadrozny2001learning} or the original isotonic regression method \cite{zadrozny2002transforming}, all predictions inside certain intervals will be mapped to be identical, and violate the injective requirement. For parametric methods, such as vector, matrix \cite{guo2017calibration}, or Dirichlet scaling \cite{kull2019beyond}, different logits can be transformed to produce the same prediction probability vectors, and violate the injective requirement.

\section{Experimental Details and Additional Results in Section~5.1}
\label{sm}
\subsection{Experimental details}
For the synthetic example, the labels ($Y$) and input features ($X$) are distributed as:
\begin{equation}
\label{generative}
\begin{split}
\mathbb{P}(Y_1=1)=\mathbb{P}(Y_2=1)=1/2;\mathbb{P}(X=x|Y_1=1) = \mathcal{N}(x;-1,1);\mathbb{P}(X=x|Y_2=1) = \mathcal{N}(x;1,1).
\end{split}
\end{equation}
The probability of observing the label $Y_1=1$ conditioned on the input $x$ can be written as: 
\begin{equation*}
\mathbb{P}(Y_1=1|X=x) = 1/[1+\exp(2x)].
\end{equation*}

We assume the prediction models to be in the following form, parameterized by $\beta_0$ and $\beta_1$:
\begin{equation*}
z=f(x)=(z_1,z_2)=\bigg(\frac{1}{1+\exp(-\beta_0-\beta_1x)},\frac{\exp(-\beta_0-\beta_1x)}{1+\exp(-\beta_0-\beta_1x)}\bigg).
\end{equation*}
This leads to close-form expressions for the canonical calibration functions $\pi(z)=(\pi_1(z),\pi_2(z))$: 
\begin{equation}
\label{mixcali}
\pi_1(z)=[1+\exp(-2\frac{\beta_0+\log(1/z_1-1)}{\beta_1})]^{-1}, \pi_2(z))=1-\pi_1(z).
\end{equation}

Finally, we estimate the ground-truth ECE$^{d}$ based on Monte Carlo integration: (i) generate $10^6$ random input-output sample pairs according to Eq.~\eqref{generative}, and (ii) record the sample average value of the quantity $|z_1-\pi_1(z)|^d$ as the ground-truth. 

\subsection{Additional results}
\begin{figure*}[h]
    \centering
    \begin{subfigure}{0.75\textwidth}{\includegraphics[width=\textwidth]{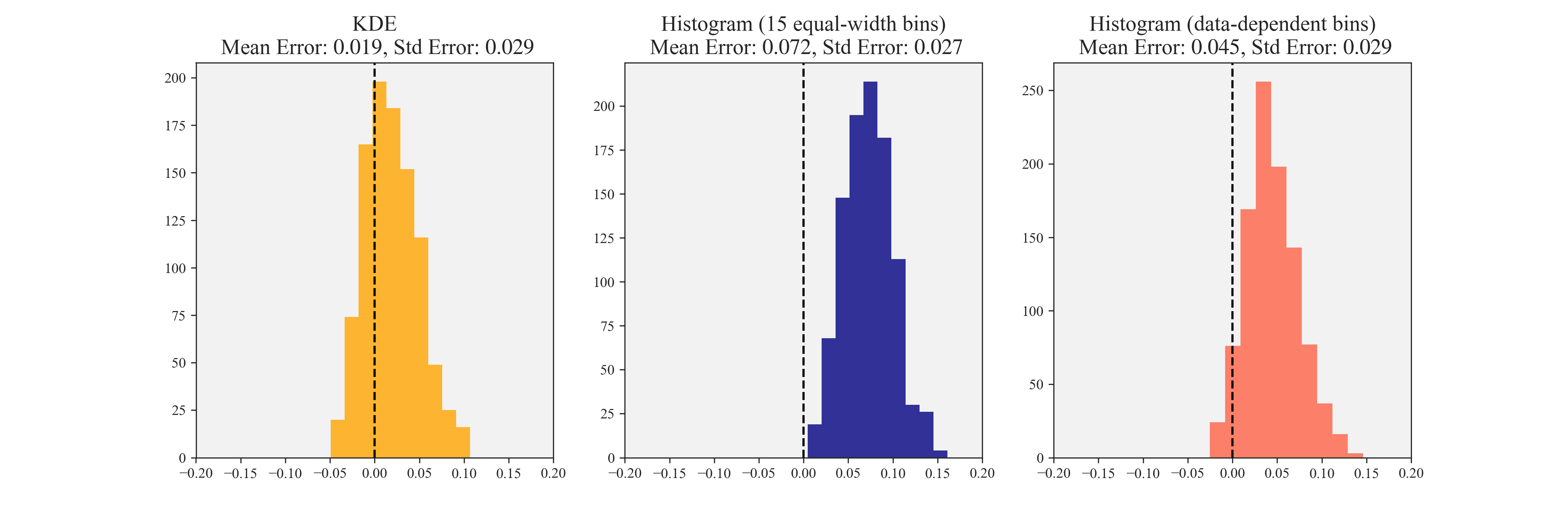}}
    \caption{$n_e=64$}	
    \end{subfigure}
    ~\\
    \begin{subfigure}{0.75\textwidth}{\includegraphics[width=\textwidth]{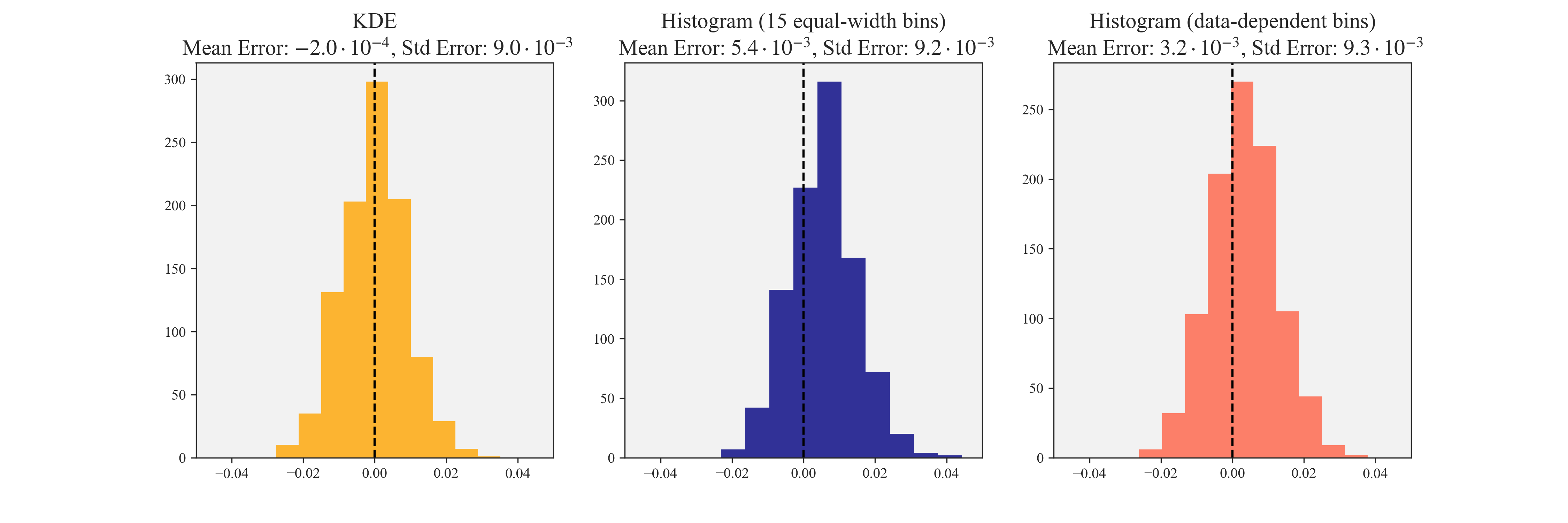}}
    \caption{$n_e=1024$}	
    \end{subfigure}   
    \caption{Distribution of ECE estimation errors in Case 1: $\beta_0=0.5,\beta_1=-1.5$ with (a) $n_e=64$ and, (b) $n_e=1024$.}
\label{f62}
\end{figure*}

\begin{figure*}[h]
    \centering
    \begin{subfigure}{0.75\textwidth}{\includegraphics[width=\textwidth]{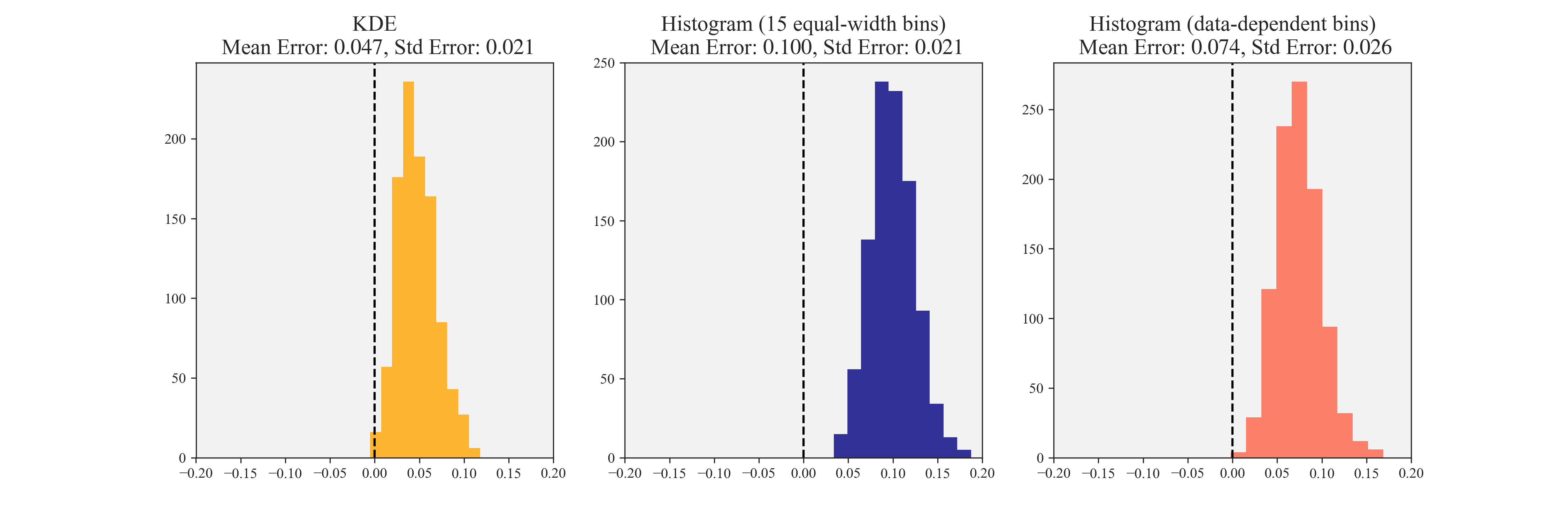}}
    \caption{$n_e=64$}	
    \end{subfigure}
    ~\\
    \begin{subfigure}{0.75\textwidth}{\includegraphics[width=\textwidth]{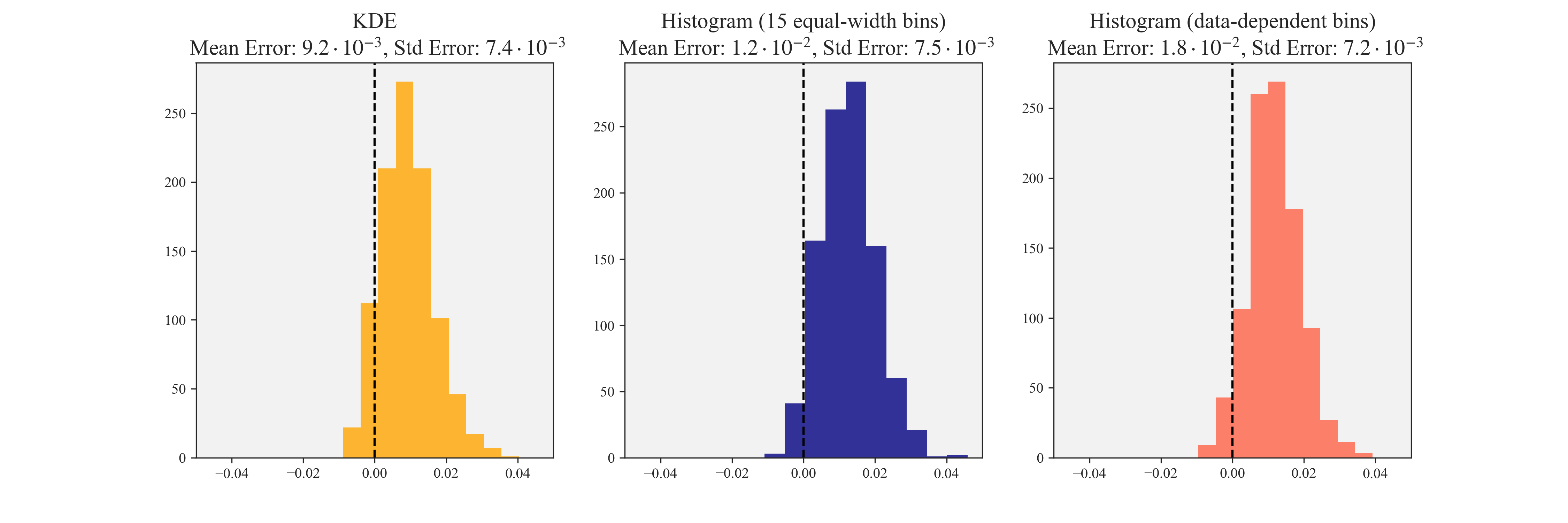}}
    \caption{$n_e=1024$}	
    \end{subfigure} 
    \caption{Distribution of ECE estimation errors in Case 2: $\beta_0=0.2,\beta_1=-1.9$ with (a) $n_e=64$, and (b) $n_e=1024$.}
\label{f63}
\end{figure*}

We plot the distribution of the errors between the ECE estimates and the ground-truth ECE in two representative scenarios: a data-limited scenario with $n_e=64$ in \autoref{f62} and a data-rich scenario with $n_e=1024$ in \autoref{f63}. The KDE estimation errors are generally less biased (more concentrated around zero) as compared to histograms, corroborating the findings in Sec.~5.1. Judging from the variance of the estimation errors, the KDE estimation errors are generally less dispersed than the two histogram estimators, indicating that the KDE estimators are more reliable. In contrast, histogram ECE estimators tend to severely over-estimates ECE in data-limited regime, with the majority of their estimation errors being positive. Their sensitivity to the binning schemes can be also observed from the distribution discrepancies between using equal-width and data-dependent bins: the histogram estimator with data-dependent bins generally performs better than the one with equal-width bins, although it cannot reach the accuracy level of KDE estimators. However, it performs the worst in the data-rich scenario of Case 2 (\autoref{f63} bottom).

\FloatBarrier

\begin{figure}[h]
\vskip 0.2in
\begin{center}
\centerline{\includegraphics[%
 width=\textwidth,clip=true]{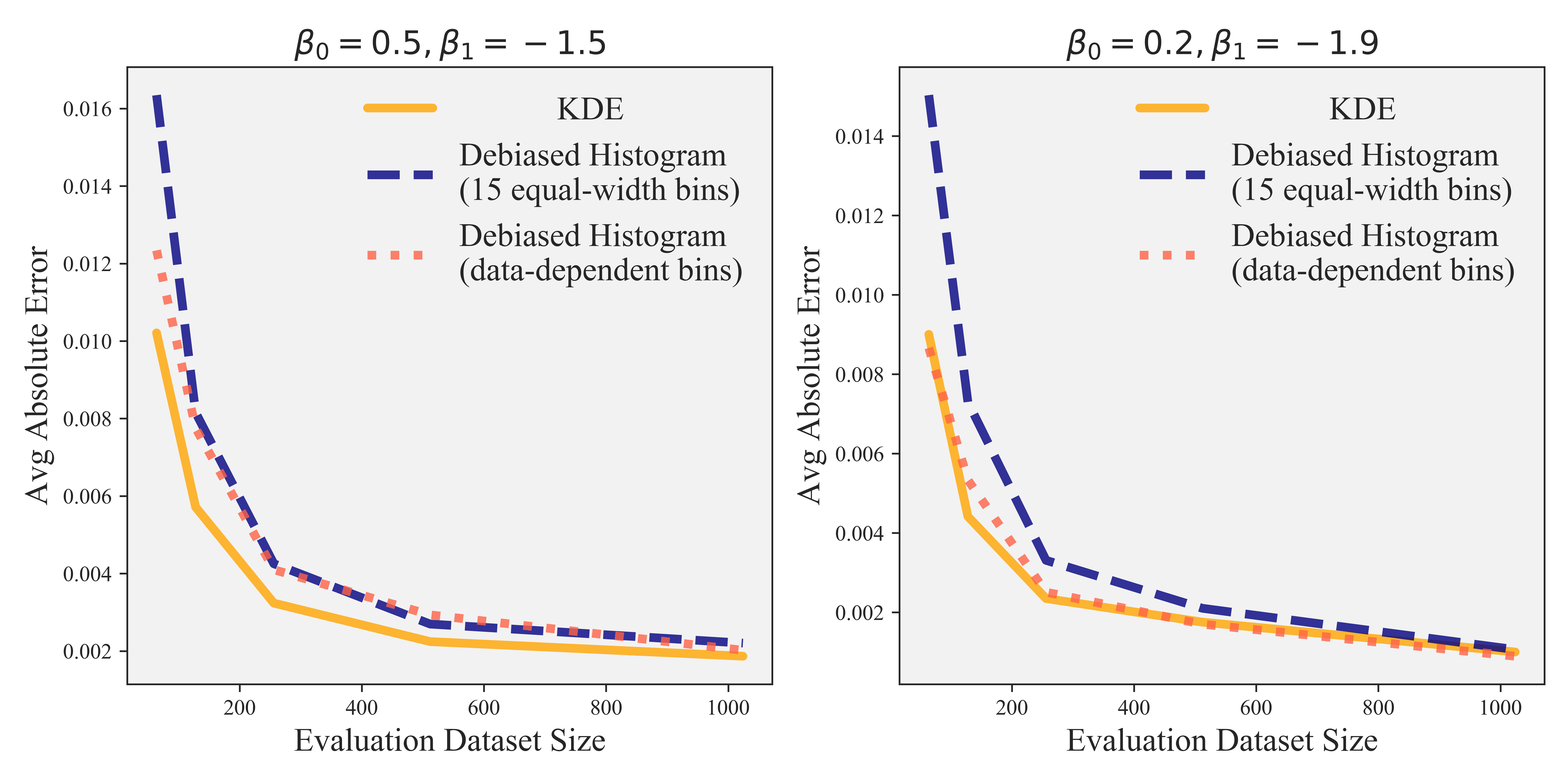}}
\caption{Average absolute error for KDE ECE and debiased histogram ECE estimators as a function of the number of samples used in the ECE estimation. KDE-based ECE estimator achieves lower estimation error, especially when the evaluation dataset is small.}
\label{debias}
\end{center}
\vskip -0.2in
\end{figure}

{Recently \cite{kumar2019verified} proposed a {\em debiased} histogram-based estimator for ECE$^{d=2}$, by leveraging error cancellations across different bins. We compare the proposed KDE ECE estimator with the debiased versions of histogram-based ECE estimators with both equal-width and data-dependent binning schemes.
We vary the size of evaluation samples $n_e$ from $64$ to $1024$ and plot the mean absolute error (averaged over $1000$ independent experiments) between KDE/Debiased-Histograms estimates for ECE$^{2}$ and the ground truth in \autoref{debias}. We observe that KDE-based estimator consistently performs better than the best-performing debiased histogram-based ECE estimators. This agrees with the findings in Sec.~\ref{sec:cee} and confirms the advantage of using KDE-based estimator over histograms-based estimators.}

\section{Experimental Detail and Additional Results in Section~5.2}
\label{sec:ar}
\subsection{Training details}
For training neural networks on CIFAR-10/100, we use SGD with Nesterov momentum and the cross-entropy loss. The weight decay is set to 0.0005, dampening to 0, momentum to 0.9, and minibatch size to 128. The initial learning rate is set to 0.1, and is dropped by a factor of 0.2 at 60, 120 and 160 epochs. For Wide ResNets, we use a dropout rate of 0.3. We train the models for a total of 500 epochs. We use standard mean/std normalization with flipping and data cropping augmentation as described in \cite{zagoruyko2016wide} on CIFAR-10/100 images. 

\subsection{Expanded results}
\begin{figure}[h]
\vskip 0.2in
\begin{center}
\centerline{\includegraphics[%
 width=\textwidth,clip=true]{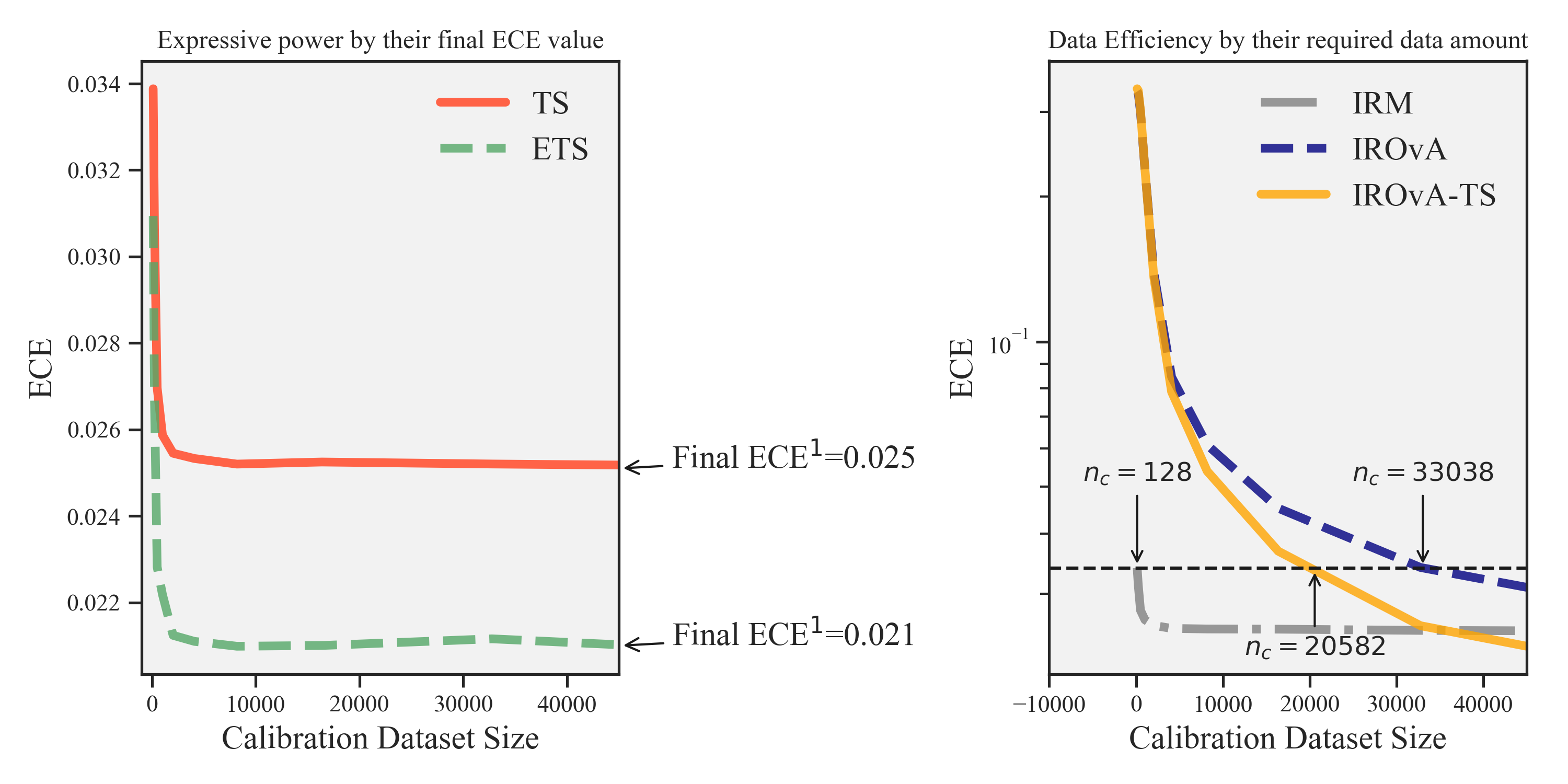}}
\caption{Graphical illustration for the proposed measure of expressive power and data-efficiency.}
\label{quant-measure}
\end{center}
\vskip -0.2in
\end{figure}

The quantitative measure of expressive power and data-efficiency is illustrated graphically in \autoref{quant-measure} and discussed in \autoref{expressive} and \autoref{efficient}. To summarize, ETS is comparably expressive to TS on CIFAR-10 and noticeably more expressive on CIFAR-100 and ImageNet. Both IRM and IROvA-TS are more efficient than IROvA. The relative efficiency gain of IRM increases as the problems become more complex. On the other hand, the relative efficiency gain of IROvA-TS appears to be quite stable on a wide range of problems. 

\begin{figure*}[h]
    \centering
    \begin{subfigure}{0.75\textwidth}{\includegraphics[width=\textwidth]{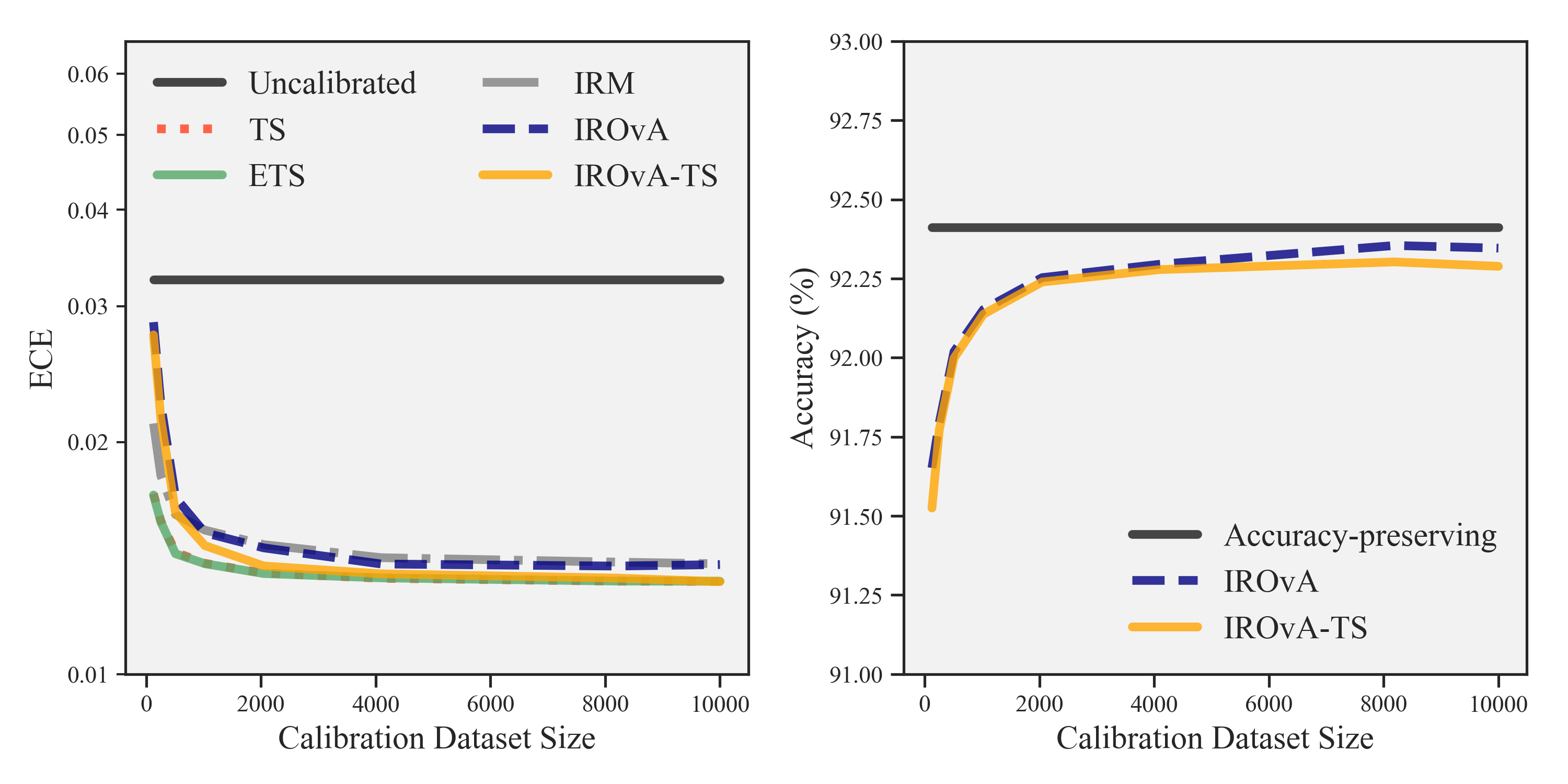}}
    \caption{CIFAR-10+DenseNet 40}	
    \end{subfigure}
    ~\\
    \begin{subfigure}{0.75\textwidth}{\includegraphics[width=\textwidth]{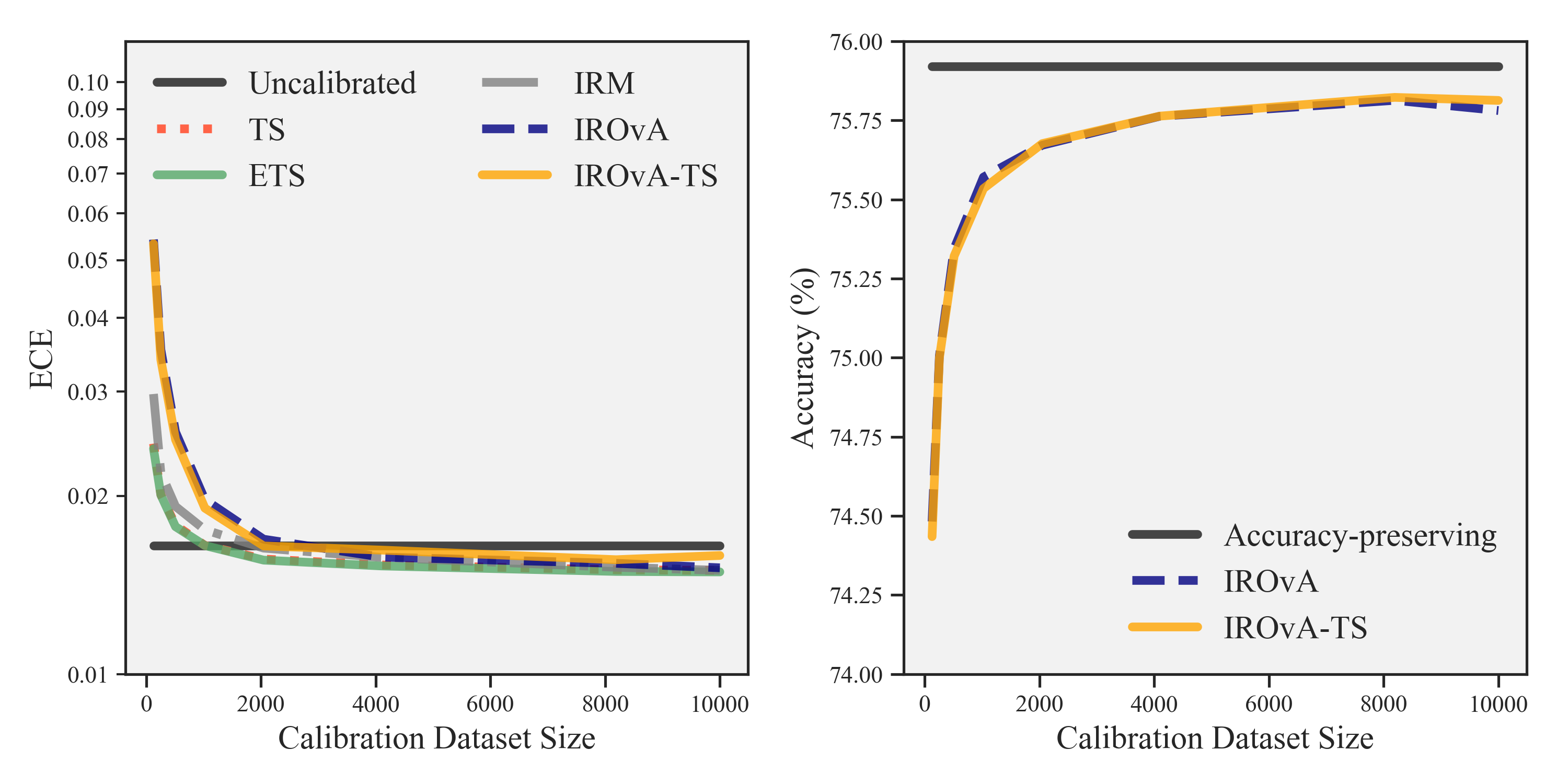}}
    \caption{CIFAR-10+LeNet 5}	
    \end{subfigure}
    ~\\
    \begin{subfigure}{0.75\textwidth}{\includegraphics[width=\textwidth]{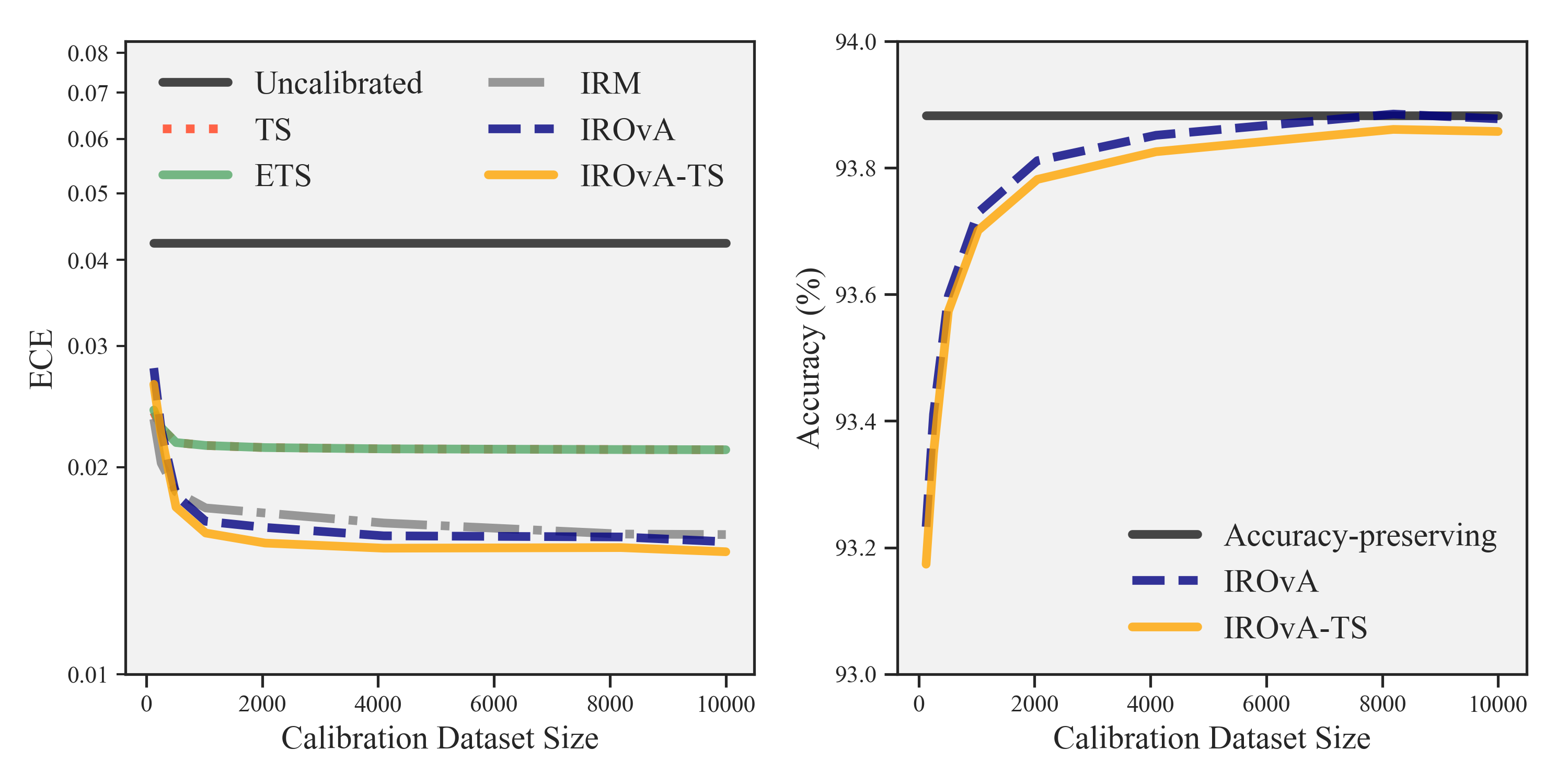}}
    \caption{CIFAR-10+ResNet 110}	
    \end{subfigure}
    \caption{Learning curve comparisons of various calibration methods on top-label ECE$^1$ (left) and the classification accuracy (right) on CIFAR-10 dataset with (a) DenseNet 40 model; (b) LeNet 5 model; and (c) ResNet 110 model.}
\label{lc-cifar10}
\end{figure*}

\begin{figure*}[h]
    \centering
    \begin{subfigure}{0.75\textwidth}{\includegraphics[width=\textwidth]{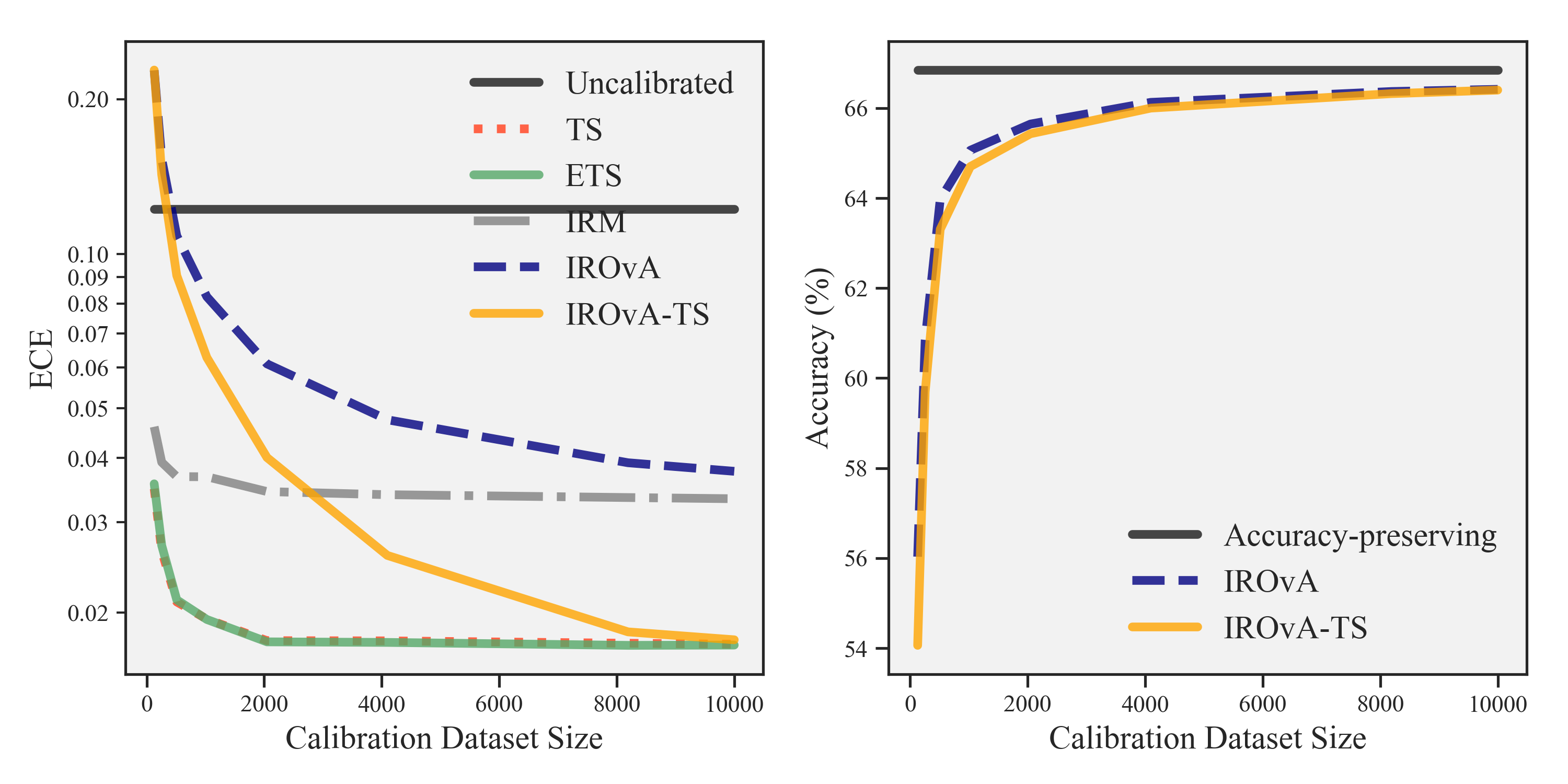}}
    \caption{CIFAR-100+DenseNet 40}	
    \end{subfigure}
    ~\\
    \begin{subfigure}{0.75\textwidth}{\includegraphics[width=\textwidth]{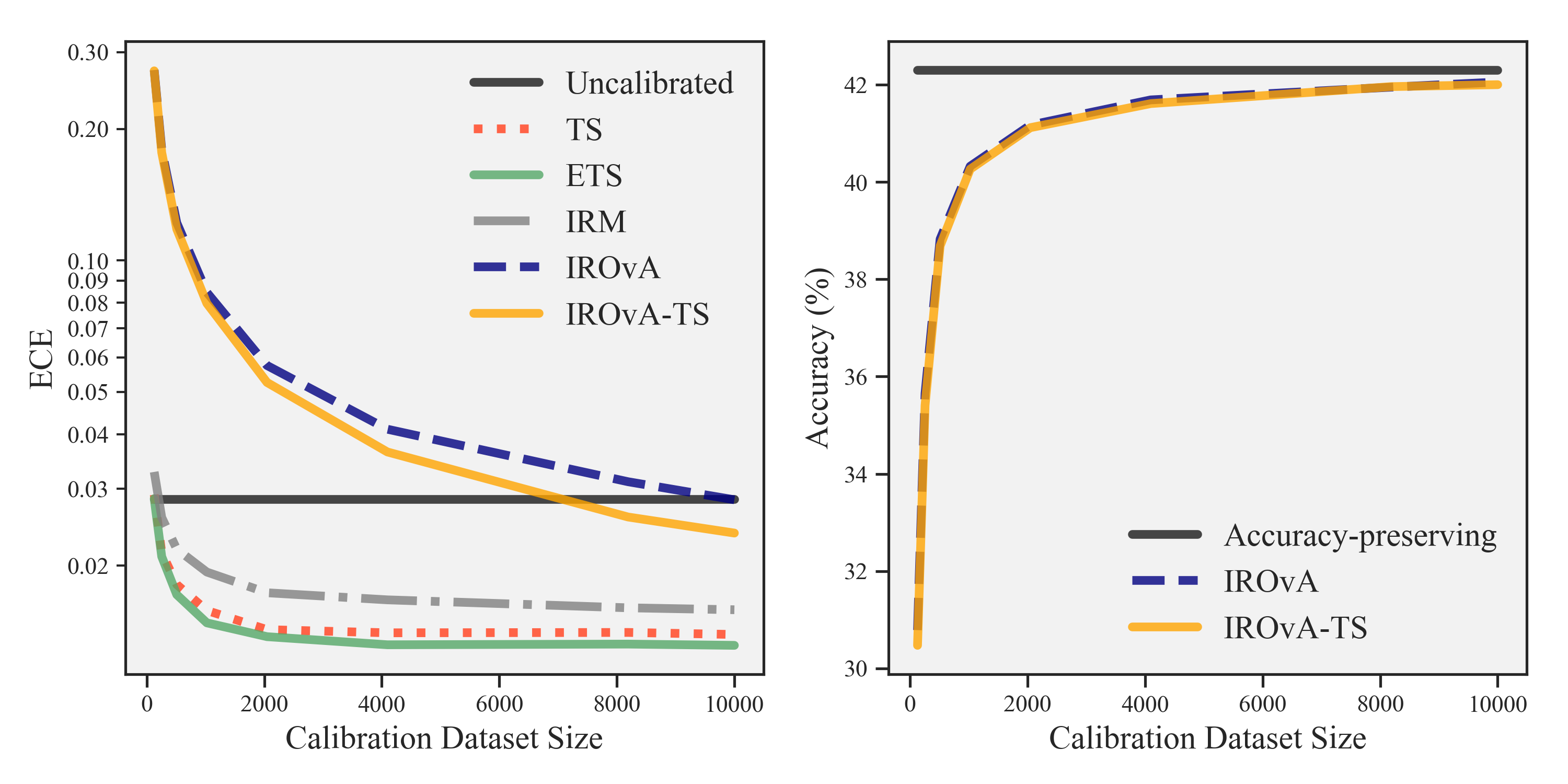}}
    \caption{CIFAR-100+LeNet 5}	
    \end{subfigure}
    ~\\
    \begin{subfigure}{0.75\textwidth}{\includegraphics[width=\textwidth]{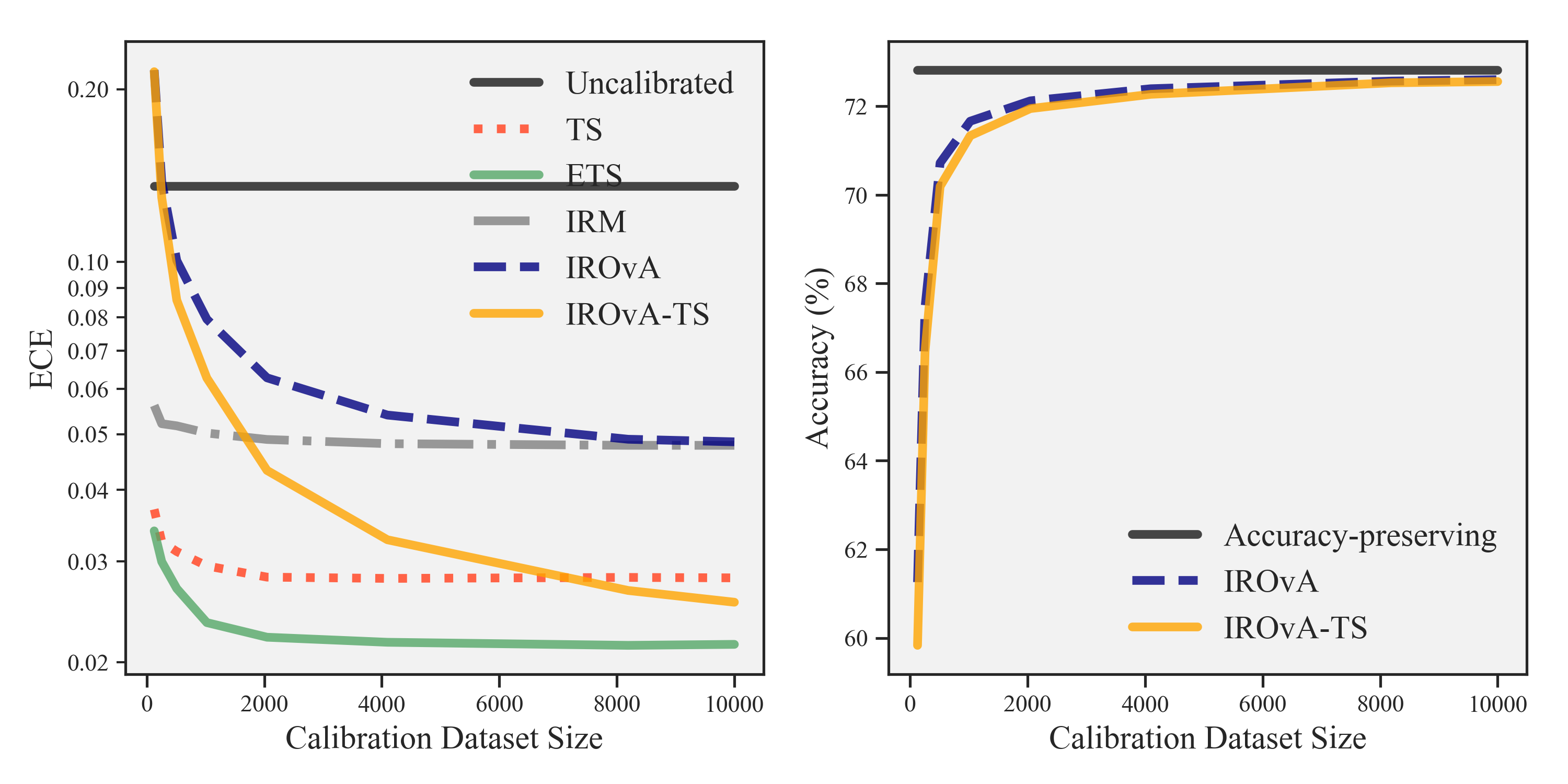}}
    \caption{CIFAR-100+ResNet 110}	
    \end{subfigure}
    \caption{Learning curve comparisons of various calibration methods on top-label ECE$^1$ (left) and the classification accuracy (right) on CIFAR-100 dataset with (a) DenseNet 40 model; (b) LeNet 5 model; and (c) ResNet 110 model.}
\label{lc-cifar100}
\end{figure*}

\begin{figure*}[h]
    \centering
    \begin{subfigure}{0.75\textwidth}{\includegraphics[width=\textwidth]{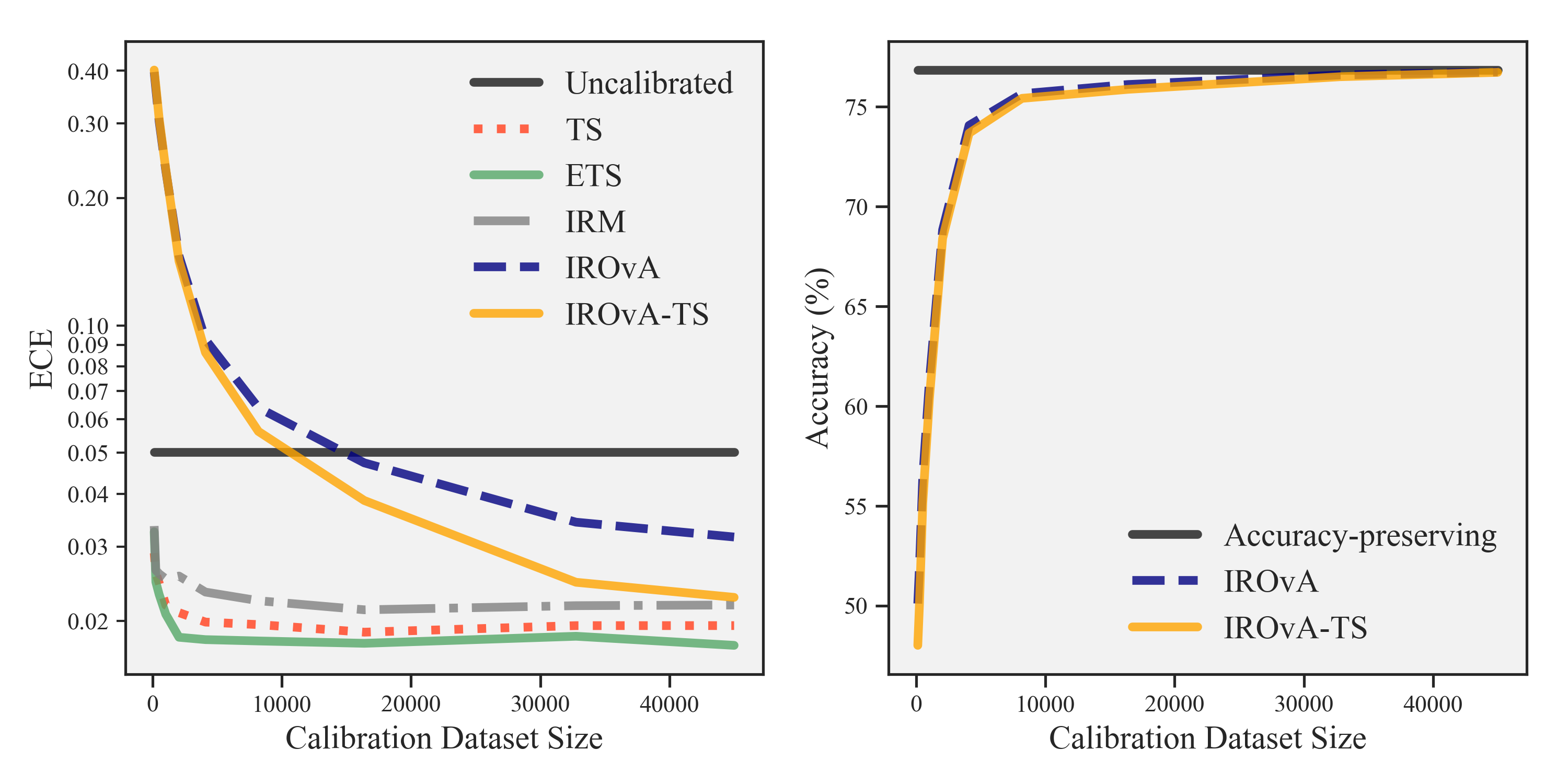}}
    \caption{ImageNet+DenseNet 161}	
    \end{subfigure}
    ~\\
    \begin{subfigure}{0.75\textwidth}{\includegraphics[width=\textwidth]{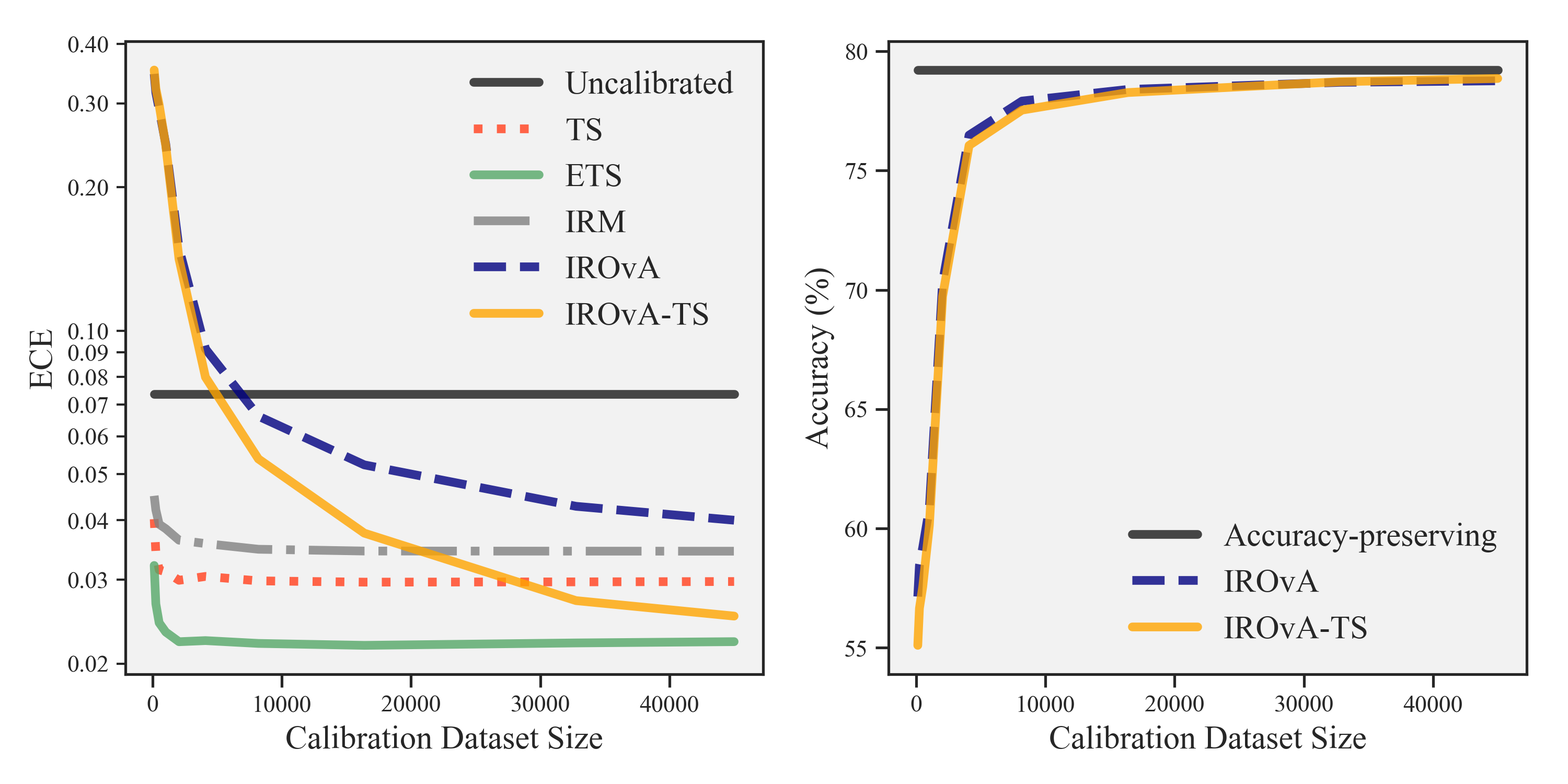}}
    \caption{ImageNet+ResNext 101}	
    \end{subfigure}
    ~\\
    \begin{subfigure}{0.75\textwidth}{\includegraphics[width=\textwidth]{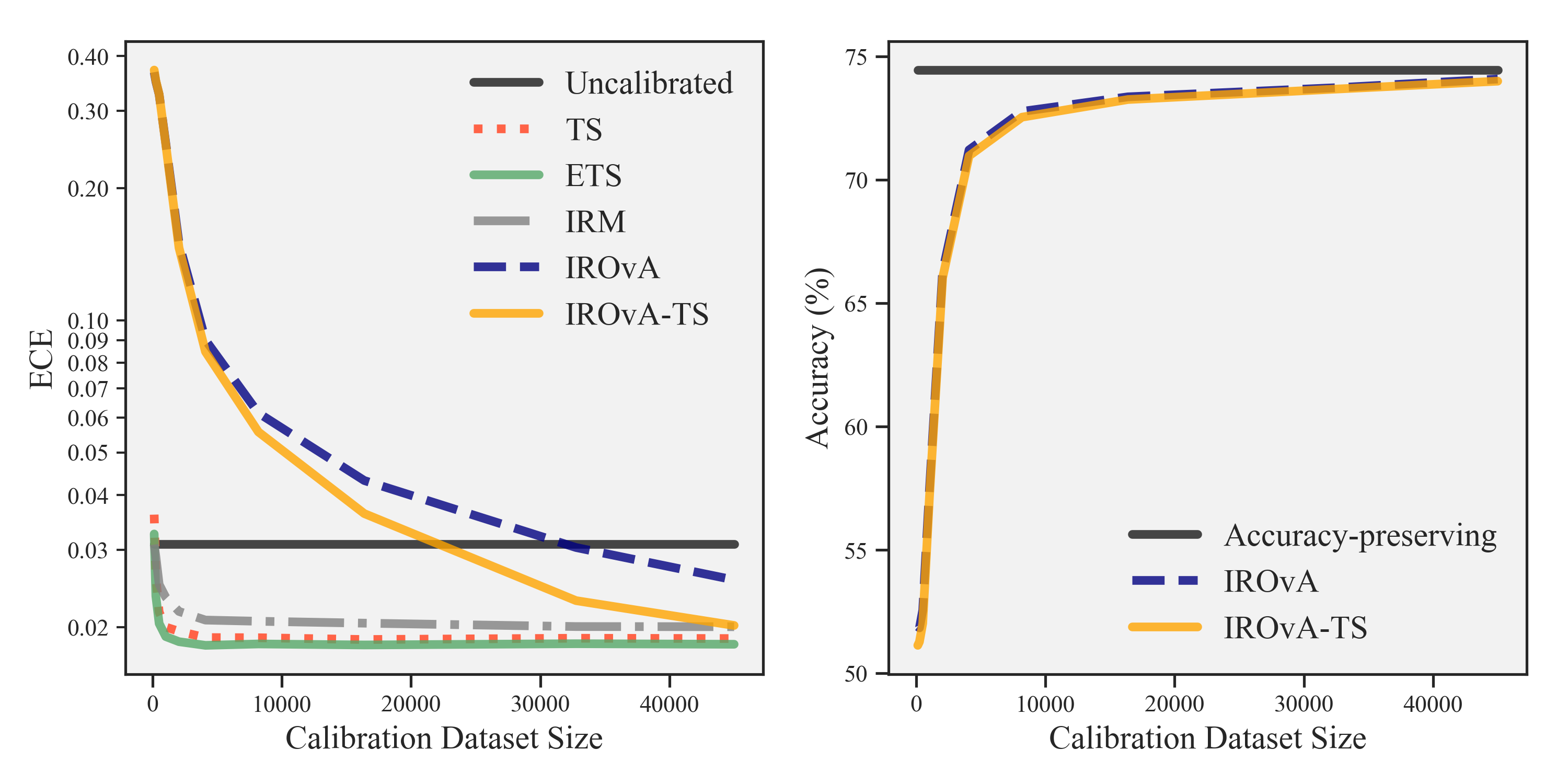}}
    \caption{ImageNet+VGG19}	
    \end{subfigure}
    \caption{Learning curve comparisons of various calibration methods on top-label ECE$^1$ (left) and the classification accuracy (right) on ImageNet dataset with (a) DenseNet 161 model; (b) ResNext 101 model; and (c) VGG 19 model.}
\label{lc-imagenet1}
\end{figure*}

We also provide expanded results on the learning curve analysis for additional neural network classifiers (see \autoref{lc-cifar10} to \autoref{lc-imagenet1}). From the visual comparison of the ECE learning curves of ETS and TS, or IRM/IROvA-TS and IROvA, we can confirm the importance to preserve the classification accuracy and the consistent benefit of employing the proposed {\it Mix-n-Match} strategies. Overall, in data-limited regime, parametric variants (TS, ETS) perform better than traditional non-parametric variants (IROvA, IROvA-TS) due to their high data-efficiency. ETS significantly outperforms TS and performs the best as the added expressive power from ensembles allow ETS to make further descent on ECE. The proposed accuracy-preserving non-parametric variant IRM also performs good: it is sometimes more effective than TS, although it cannot outperform ETS in most examined cases. The relatively good performance of IRM can be accredited to its high data-efficiency on complex calibration tasks. Going to the data-rich regime, the ECE reduction progress stalls for parametric methods. On complex problems, such as CIFAR-100 and ImageNet, this also applies to the accuracy-preserving non-parametric variants (IRM) due to its expressivity-efficiency trade-off. In contrast, the high expressive power of non-parametric variants (IROvA, IROvA-TS) allow them to keep minimizing the ECE and eventually outperform less expressive methods (ETS, TS or IRM) with sufficient amount of data -- although, this cannot be verified in all the examined cases due to our limited data budget. In such regime, the composition method IROvA-TS significantly outperforms IROvA and performs the best due to its enhanced data-efficiency. Based on such observations, our further discussion on the guidelines of calibration methods will be restricted to ETS, IRM and IROvA-TS. 

\begin{table}[h]
\caption{ECE$^{1}$ (\%) with $n_c=10000$ for CIFAR and $n_c=45000$ for ImageNet; lower values imply more expressive power.}
\label{expressive}
\vskip 0.15in
\begin{center}
\begin{small}
       \begin{tabular}{llcc}
        \toprule
   Dataset & Model    & TS & ETS  \\
        \midrule
        CIFAR-10 & DenseNet 40 & 1.32  & 1.32\\
        CIFAR-10 & LeNet 5 & 1.49  & \textbf{1.48}\\
        CIFAR-10 & ResNet 110 & 2.12  & 2.12\\
        CIFAR-10 & WRN 28-10 & 1.67  & 1.67 \\
        \midrule
        CIFAR-100 &DenseNet 40 & 1.73  & \textbf{1.72}\\
        CIFAR-100 &LeNet 5 & 1.39  & \textbf{1.31}\\
        CIFAR-100 &ResNet 110 & 2.81  & \textbf{2.15}\\
        CIFAR-100 &WRN 28-10 & 3.23  & \textbf{2.85}\\
        \midrule
        ImageNet & DenseNet 161 & 1.95  & \textbf{1.75}\\
        ImageNet & ResNeXt 101 & 2.97  & \textbf{2.22}\\
        ImageNet & VGG 19 & 1.89  & \textbf{1.83}\\
        ImageNet & WRN 50-2 & 2.52  & \textbf{2.10}\\
        \bottomrule
        \end{tabular}
\end{small}
\end{center}
\vskip -0.1in
\end{table}

\begin{table}[h]
\caption{Required calibration data amount $n_c$ to reach IRM's performance with $n_c=128$ samples; lower value means more data-efficient. All values are normalized (divided by 128) to show how many samples are equivalent to one sample in IRM for each method.}
\label{efficient}
\vskip 0.15in
\begin{center}
\begin{small}
       \begin{tabular}{llccc}
        \toprule
        Dataset &  Model & IRM & IROvA & IROvA-TS  \\
        \midrule
       CIFAR-10 & DenseNet 40 & \textbf{1.0}  & 2.45 & 2.10 \\
       CIFAR-10 & LeNet 5 & \textbf{1.0}  & 3.15 & 2.92\\
       CIFAR-10 & ResNet 110 & \textbf{1.0}  & 1.84 & 1.70\\
       CIFAR-10 & WRN 28-10 & \textbf{1.0}  & 1.98 & 1.90\\
        \midrule
     CIFAR-100 & DenseNet 40 & \textbf{1.0}  & 37.6 & 17.9\\
     CIFAR-100 &   LeNet 5 & \textbf{1.0}  & 58.7 & 43.0 \\
     CIFAR-100 &   ResNet 110 & \textbf{1.0}  & 28.1 & 10.7\\
     CIFAR-100 &   WRN 28-10 & \textbf{1.0}  & 24.2 & 15.6\\
        \midrule
     ImageNet & DenseNet 161 & \textbf{1.0}  & 282 & 174\\
     ImageNet & ResNeXt 101 & \textbf{1.0}  & 226 & 98.6\\
     ImageNet &   VGG 19 & \textbf{1.0}  & 251 & 180\\
     ImageNet &   WRN 50-2 & \textbf{1.0}  & 258 & 161\\
        \bottomrule
        \end{tabular}
\end{small}
\end{center}
\vskip -0.1in
\end{table}

\FloatBarrier
\section{Guidelines}
\label{sec:guide}

First, we provide guidelines on choosing an appropriate calibration evaluation metric. If knowing the exact value of ECE is important, we recommend KDE-based top-label ECE estimator for its superior data-efficiency as compared to histograms. 
If the goal is to infer just the rankings not actual calibration errors, one should use the calibration gain metric. It provides a reliable and faithful comparison of different methods based on their actual calibration capabilities. We also note that calibration gain metric might be a lower bound for certain calibration methods, e.g., non accuracy-preserving methods. 

We next provide general guidelines on selecting the best calibration method (ETS vs. IRM vs. IROvA-TS), based on: (a) the complexity of the calibration task which is a function of the model complexity (number of free parameters) and the data complexity (number of classes), and (b) resources at hand (the amount of the calibration data).

The complexity of the calibration task is directly related to the complexity of the canonical calibration function in Eq.~\eqref{pi}. Although, we do not have the knowledge of the canonical calibration function, we expect a learning task with low model complexity and low data complexity to result in a low complexity calibration task ( see \autoref{lc-cifar10} (b)).
Next, a learning task with low model complexity but high data complexity (\autoref{lc-cifar100} (b)), or high model complexity but low data complexity (\autoref{lc-cifar10} (a) and (c)) is expected to result in a moderately complex calibration task. Finally, we expect a learning task with high model \& data complexity to result in a highly complex calibration task (\autoref{lc-cifar100} (a) and (c), \autoref{lc-imagenet1}).

For low complexity calibration tasks, we see that the performance of uncalibrated models are already satisfactory. This observation agrees with results in~\cite{guo2017calibration}. Further, all the calibration methods perform similarly, however, proposed variants performing slightly better than the baseline approaches. The use case of the most practical interest is where the calibration task is expected to be complex. In such scenarios, an ideal calibration map should have enough expressive power to accurately approximate the canonical calibration function in Eq.~\eqref{pi}. However, to fit an expressive calibration map, sufficiently large amount of calibration data is required which may or may not be available. In data limited regime, ETS is recommended as the first choice, while IRM is a potential alternative when the parametric assumptions of ETS are improper (see \autoref{lcimagenet} top left and \autoref{lc-cifar10} (c)). In data rich regime, we recommend using IROvA-TS for its high expressive power. For moderate complexity calibration tasks, the patterns are similar to high complexity calibration tasks. The only difference is that the gain of the proposed {\it Mix-n-Match} strategies are not as drastic as of the case where calibration task is of high complexity. 
Of course, if the user has hard-constraints on accuracy-preservation, the choice would be limited to the accuracy-preserving calibrators regardless of the data size or the task complexity. In such scenarios, we recommend ETS and IRM. We also want to emphasize that both ETS and IRM are fairly efficient and perform well on all ranges of the calibration data size. 

In summary, our take-home messages on the most appropriate calibration method are the following:
\begin{itemize}
    \item For complex calibration task (poorly calibrated model, large number of classes), when a large calibration dataset is available and the user does not have hard constraints on preserving accuracy, the proposed compositional method IROvA-TS is recommended to achieve the best degree of calibration.
    
    \item For all other cases, the proposed ensemble method ETS is recommended when the parametric assumption is proper. IRM is a strong alternative to be considered in order to avoid the risk of parametric form mis-specification of ETS in certain cases. Both approaches preserve the classification accuracy, and one can conveniently compare them based on the proposed calibration gain metric.
\end{itemize}

\end{appendices}	

\end{document}